%% file: main.tex
\documentclass{article} 
\usepackage{iclr2027_conference,times}
\usepackage{graphicx}
\usepackage{booktabs}
\usepackage{multirow}
\usepackage{colortbl}
\usepackage{wrapfig}
\usepackage{needspace}

\input{math_commands.tex}

\usepackage{hyperref}
\usepackage{url}

\title{In-Context Preference-Conditioned Pareto Set Learning from Limited Offline Data}

\author{
Yiming Yao\textsuperscript{1},
Liang Zhao\textsuperscript{1} \&
Qingfu Zhang\textsuperscript{1}\\
{\normalfont\textsuperscript{1}City University of Hong Kong}\\
{\normalfont\small
\texttt{yimingyao3-c@my.cityu.edu.hk},
\texttt{liazhao5-c@my.cityu.edu.hk}}\\
{\normalfont\small\texttt{qingfu.zhang@cityu.edu.hk}}
}

\iclrfinalcopy
\begin{document}

\maketitle
\lhead{Preprint}

\input{sections/00_abstract.tex}
\input{sections/01_introduction.tex}
\input{sections/02_related_work.tex}
\input{sections/03_methodology.tex}
\input{sections/04_experiments.tex}
\input{sections/05_conclusion.tex}

\input{sections/iclr_ai_use_statement.tex}
\input{sections/iclr_reproducibility_statement.tex}

\bibliography{iclr2027_conference}
\bibliographystyle{iclr2027_conference}

\input{sections/06_appendix.tex}

\end{document}

%% file: math_commands.tex
\usepackage{amsmath,amsfonts,bm}

\def\eqref#1{equation~\ref{#1}}

\def\1{\bm{1}}

\DeclareMathAlphabet{\mathsfit}{\encodingdefault}{\sfdefault}{m}{sl}
\SetMathAlphabet{\mathsfit}{bold}{\encodingdefault}{\sfdefault}{bx}{n}



%% file: sections/00_abstract.tex
\begin{abstract}

Multi-objective optimization involves diverse preferences over conflicting objectives. Rather than returning a finite set of trade-off solutions, preference-conditioned Pareto set learning seeks to map a specified preference directly to a corresponding Pareto optimal solution. Yet in many real-world scenarios, optimization must proceed offline without further objective evaluations, leaving only a limited static dataset that sparsely covers the decision and objective spaces. Existing methods either optimize separately for each preference or train a problem-specific Pareto solution model, requiring repeated search across preferences or repeated training across problems. We introduce in-context preference-conditioned Pareto set learning (ICPSL), a framework that amortizes both preference-wise solution search and problem-specific model construction. ICPSL pretrains a foundation-style hypernetwork on a large and diverse set of synthetic prior data. Given the offline observations of an unseen problem, the pretrained hypernetwork learns to generate a problem-specific, preference-conditioned Pareto solution model in a single forward pass. The generated model can then be reused to predict solutions for arbitrary preferences without problem-specific training or preference-wise search. Experiments on synthetic and real-world problems show that ICPSL produces high-quality Pareto solutions while providing substantial speedups in problem-specific solution-model construction.

\end{abstract}

%% file: sections/01_introduction.tex
\section{Introduction}
\label{sec:introduction}

Many real-world optimization problems involve multiple conflicting objectives, where improving one objective may necessarily compromise another~\citep{ying2019bench,tanabe2020easy,dara2022machine}. Consequently, there is generally no single optimal solution, the goal is to identify a Pareto set that characterizes the possible trade-offs among objectives. Conventional methods approximate this set with a finite collection of solutions, which may not include the solution that best matches a decision-maker's preferred trade-off~\citep{deb2002fast,beume2007sms}. Preference-conditioned Pareto set learning (PSL) instead represents the trade-off structure as a model that maps an arbitrary preference $\boldsymbol{\lambda}$ to the corresponding optimal solution $\boldsymbol{x}^{\star}(\boldsymbol{\lambda})$~\citep{navon2021learning,lin2022pareto}. By varying the specified preference, decision-makers can readily explore different trade-off regions of the approximate Pareto set, enabling flexible decision-making.

However, in many practical scenarios, this task becomes more challenging when optimization must proceed offline using only a limited, fixed dataset of previous observations without further queries to the true evaluations~\citep{trabucco2021conservative,yu2021roma,chen2023parallel}. As such datasets typically cover only a small portion of the decision and objective spaces, they provide limited information about promising designs for query preferences~\citep{xue2024offline,yuan2025paretoflow,zhu2026domoo}. The key challenge is therefore how to use the transferable and integrable information in the limited observations to infer high-quality Pareto optimal solutions for arbitrary unseen preferences.

Existing methods follow two main paradigms for producing preference-conditioned solutions. The distinction is whether the predicted solution $\hat{\boldsymbol{x}}(\boldsymbol{\lambda})$ is induced implicitly or represented explicitly as a solution model. Implicit method such as ParetoFlow learns a flow model of offline decisions and objective predictors, then uses the predictors to guide generation for all queried preferences~\citep{yuan2025paretoflow}. This paradigm focuses how well each preference-conditioned solution can be optimized given a reliable instance-specific model. However, this procedure still requires test-time search for each query preference, and makes the final solution sensitive to surrogate error when data are limited. By contrast, explicit methods exploit regularity across preferences to directly learn a solution model~\citep{navon2021learning}, typically a multilayer perceptron (MLP), that approximates the preference-to-solution mapping $\boldsymbol{\lambda}\mapsto \hat{\boldsymbol{x}}(\boldsymbol{\lambda})\approx\boldsymbol{x}^\star(\boldsymbol{\lambda})$. Representative approaches include regression-based Pareto set learning (PSLR), developed from regression-based amortized optimization~\citep{amos2023tutorial}, as well as classic objective-based Pareto set learning (PSLO)~\citep{lin2022pareto} and the recently proposed DOMOO~\citep{zhu2026domoo}. Although these methods avoid repeated test-time search across preferences, they typically require training a solution model from scratch for each new problem instance. This motivates a general mechanism for explicitly learning a high-quality solution model from a limited offline static dataset, while avoiding both per-problem retraining and per-preference test-time search.

To this end, we propose \emph{in-context preference-conditioned Pareto set learning (ICPSL)}. The core component is a Transformer-based, foundation-style hypernetwork pretrained on extensive synthetic prior data. Given the offline observations of a new problem instance, the hypernetwork performs in-context inference and directly generates an problem-specific solution model in a single forward pass. The generated MLP solution model explicitly represents the preference-to-solution mapping and returns a solution for any query preference through a single MLP forward pass. The framework therefore realizes two levels of amortization: the generated solution model amortizes solution search across query preferences for a given problem, while the pretrained hypernetwork amortizes solution-model construction across various problem instances. In this view, ICPSL moves beyond per-preference solution search and per-problem model training, shifting toward a foundation-style paradigm in which a pretrained model learns to use offline observations in context to construct solution models for arbitrary query preferences.

Our contributions are summarized as follows:
\begin{itemize}
\item We introduce ICPSL, a framework in which a foundation-style hypernetwork is pretrained on extensive synthetic prior data to generate a reusable problem-specific solution model from limited offline observations in one forward pass, thereby amortizing both solution search across preferences and model construction across problems.
\item To accommodate potential discontinuities in the optimal preference-to-solution mapping, we model a preference-conditioned solution distribution through latent-conditioned components and develop three inference strategies to predict solutions.
\item Compared with representative baselines, the proposed framework produces high-quality Pareto solutions on synthetic problems and real-world engineering applications, while providing substantial speedups in problem-specific solution-model construction.
\end{itemize}

%% file: sections/02_related_work.tex
\section{Preliminaries and Related Work}
\label{sec:preliminaries-related-work}

\subsection{Preliminaries}

\paragraph{Problem Definition.}In this paper, we consider the following multi-objective optimization problem (MOP):
\begin{equation}
\min _{\boldsymbol{x} \in \mathcal{X}} \boldsymbol{f}(\boldsymbol{x})=(f_1(\boldsymbol{x}), f_2(\boldsymbol{x}), \cdots, f_m(\boldsymbol{x})),
\end{equation}
where $\boldsymbol{x}=(x_{1}, x_{2}, \ldots, x_{d})^T$ is the feature vector in the decision space $\mathcal{X} \subset \mathbb{R}^d$, and $\boldsymbol{f}: \mathcal{X} \rightarrow \mathbb{R}^m$ is the objective vector containing $m$ continuous objective functions. Typically, there is no single solution that can simultaneously achieve the optimal for all objectives. Instead, the goal is to find a set of solutions with Pareto optimal objective trade-offs. We say an objective vector $\boldsymbol{f}(\boldsymbol{x})$ 
Pareto dominates another $\boldsymbol{f}(\boldsymbol{x}^{\prime})$, denoted as $\boldsymbol{f}(\boldsymbol{x})\prec\boldsymbol{f}(\boldsymbol{x}^{\prime})$ if $f_{i}(\boldsymbol{x}) \leq f_{i}(\boldsymbol{x}^{\prime}), \forall i\in\{1, \ldots, m\}$ and there exists $j \in\{1, \ldots, m\}$ such that $ f_{j}(\boldsymbol{x}) < f_{j}(\boldsymbol{x}^{\prime})$. A solution $\boldsymbol{x}^*\in\mathcal{X}$ is Pareto optimal if there is no $\boldsymbol{x}$ such that $\boldsymbol{f}(\boldsymbol{x})\prec \boldsymbol{f}(\boldsymbol{x}^*)$. The set of all Pareto optimal solutions is called the Pareto set (PS), and the image of PS in the objective space is called the Pareto front (PF).

\paragraph{Tchebycheff Scalarization.} Given the preference vector $\boldsymbol{\lambda}\in\Delta^{m-1}$, where $\Delta^{m-1}=\{\boldsymbol{\lambda}\in\mathbb{R}_{+}^{m}: \sum_{j=1}^{m}\lambda_j=1\}$, Tchebycheff scalarization measures the largest weighted deviation from the ideal objective vector $\boldsymbol{f}^{\mathrm{ideal}}$, with $f_j^{\mathrm{ideal}}=\inf_{\boldsymbol{x}\in\mathcal X}f_j(\boldsymbol{x})$:
\begin{equation}
s_{\boldsymbol{\lambda}}(\boldsymbol{x})
=\max_{j\in\{1,\ldots,m\}}
\lambda_j\bigl(f_j(\boldsymbol{x})-f_j^{\mathrm{ideal}}\bigr).
\label{eq:prelim-tchebycheff-scalarization}
\end{equation}
With a suitable ideal point, each Pareto optimal solution can be obtained by optimizing the Tchebycheff scalarization with a specific but unknown preference~\citep{choo1983proper}:
\begin{equation}
    \boldsymbol{x}^{\star}(\boldsymbol{\lambda})
    = \arg\min_{\boldsymbol{x}\in\mathcal{X}}
    s_{\boldsymbol{\lambda}}(\boldsymbol{x}).
\end{equation}
We study the problem when the objectives are unknown and the sole source of information is a fixed offline dataset $\mathcal{D}=\{(\boldsymbol{x}_i,\boldsymbol{y}_i)\}_{i=1}^{N_c}$, with $\boldsymbol{y}_i=\boldsymbol{f}(\boldsymbol{x}_i)$. No additional objective evaluation is allowed during training or inference. Our goal is to use $\mathcal{D}$ to infer all preference-conditioned Pareto optimal approximates $\hat{\boldsymbol{x}}(\boldsymbol{\lambda})\approx\boldsymbol{x}^\star(\boldsymbol{\lambda})$ for arbitrary preferences.

\subsection{Related Work}

\paragraph{Pareto Set Learning.}
Pareto set learning (PSL) amortizes solution search across preferences by learning a preference-to-solution mapping for a given problem. Pareto HyperNetworks use a hypernetwork to generate preference-conditioned target-network weights~\citep{navon2021learning}, while PSL-MOBO applies preference-conditioned modeling to expensive black-box optimization~\citep{lin2022pareto}. Subsequent work addresses different challenges in learning this mapping. EPS adapts preference sampling to the shape of the Pareto front~\citep{ye2024evolutionary}, whereas CDM-PSL uses diffusion models to capture complex solution distributions in multi-objective Bayesian optimization~\citep{li2025expensive}. These advances improve preference coverage or solution generation within a target problem, but still rely on problem-specific training rather than constructing a reusable model directly from a fixed offline context.

\paragraph{Offline Multi-objective Optimization.}
Offline model-based optimization seeks improved designs from fixed observations without further objective evaluations. Earlier work addresses challenges of this setting through conservative modeling, robust adaptation, and parallel mentoring~\citep{trabucco2021conservative,yu2021roma,chen2023parallel}. The benchmark of \citet{xue2024offline} extends this setting to multiple objectives. Within offline MOO, ParetoFlow guides samples from a learned flow using objective predictors~\citep{yuan2025paretoflow}, whereas DOMOO combines objective surrogates and out-of-distribution risk modeling with nested PSL~\citep{zhu2026domoo}. These methods respectively improve preference-conditioned generation and mitigate out-of-distribution extrapolation, but still rely on problem-specific training or preference-wise search rather than constructing a reusable solution model in one forward pass.

\paragraph{Amortized Multi-objective Optimization.}
Amortized optimization learns to predict solutions across related problem instances instead of solving each instance from scratch~\citep{amos2023tutorial}. Recent work extends this principle to MOO through pretrained foundation-style models that transfer across problem instances. BOFormer uses Transformer and non-Markovian reinforcement learning to learn a multi-objective Bayesian-optimization policy~\citep{hung2025boformer}, while TAMO pretrains a task-agnostic policy that proposes the next design from an optimization history in a single forward pass~\citep{zhang2026tamo}. Both methods amortize candidate generation, but operate in an online setting in which proposed designs receive further objective evaluations. Moreover, they do not provide preference-conditioned solutions and thus cannot explore different trade-off regions by adjusting queried preferences.

%% file: sections/03_methodology.tex
\section{Methodology}
\label{sec:methodology}

\subsection{Framework Overview}
\label{sec:framework-overview}

\begin{figure*}[t]
\centering
\includegraphics[width=\textwidth]{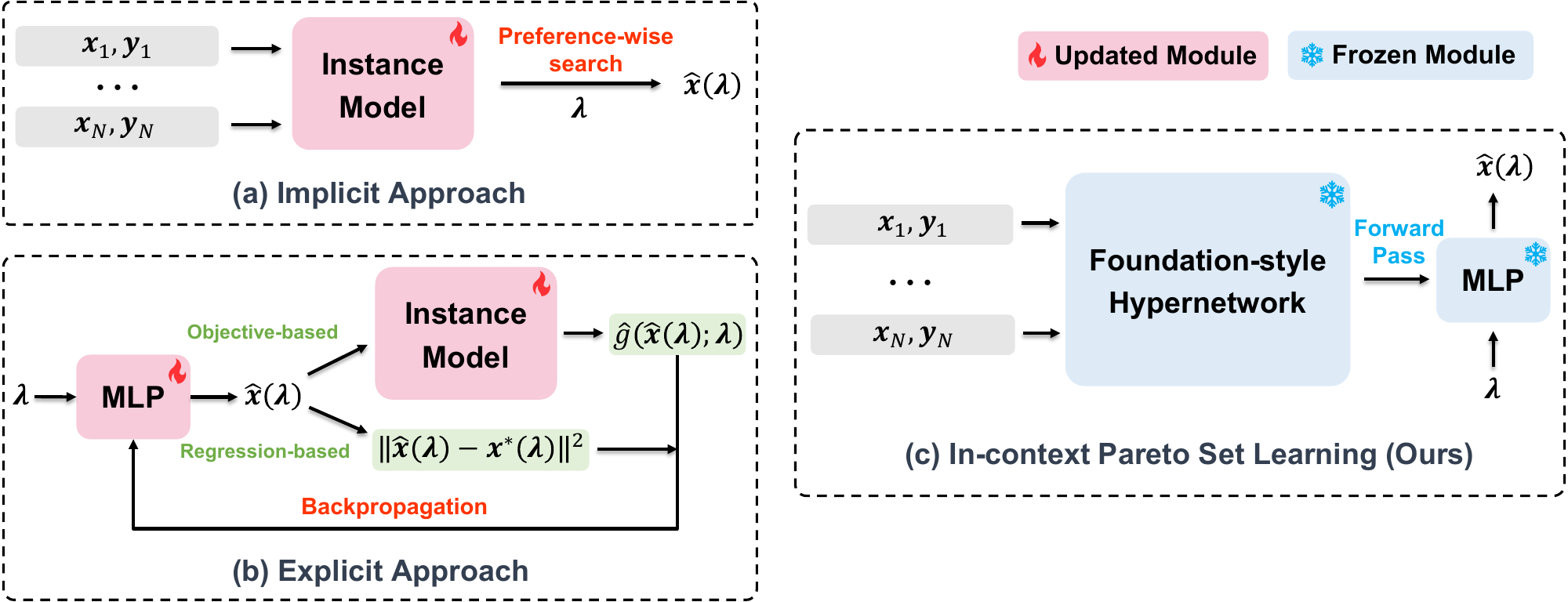}
\caption{Comparison of offline preference-conditioned Pareto set learning paradigms. (a) The implicit approach performs preference-wise search over a problem-specific instance model. (b) The explicit approach trains a problem-specific MLP as the Pareto solution model through objective- or regression-based learning. (c) ICPSL uses a frozen foundation-style hypernetwork to generate an problem-specific MLP from offline observations in a single forward pass, which maps arbitrary preferences to solutions without problem-specific training or preference-wise search.}
\label{fig:framework-overview}
\end{figure*}

As shown in Figure~\ref{fig:framework-overview}, ICPSL learns to directly construct a problem-specific Pareto solution model from the offline observations $\mathcal{D}$ of a new multi-objective optimization problem. To this end, we pretrain a foundation-style hypernetwork $H_{\boldsymbol{\theta}}$ on a large and diverse collection of synthetic multi-objective problems. At inference, the pretrained hypernetwork remains fixed and maps the offline observations $\mathcal{D}$ from an unseen target problem to the parameters $\boldsymbol{\phi}_{\mathcal{D}}=H_{\boldsymbol{\theta}}(\mathcal{D})$ of a problem-specific Pareto solution model.  The resulting model $h_{\boldsymbol{\phi}_{\mathcal{D}}}$ is a lightweight MLP instantiated directly from the hypernetwork output. Without separate training or fine-tuning, it serves as a reusable Pareto solution model that maps an arbitrary preference $\boldsymbol{\lambda}$ to the corresponding predicted solution as $\hat{\boldsymbol{x}}_{\mathcal{D}}(\boldsymbol{\lambda})=h_{\boldsymbol{\phi}_{\mathcal{D}}}(\boldsymbol{\lambda})$.

Consequently, ICPSL requires neither problem-specific training nor test-time optimization over the queried preferences. This efficiency stems from two complementary levels of amortization. The pretrained hypernetwork first amortizes Pareto solution-model construction across problem instances, while the generated MLP subsequently amortizes solution search across preferences within each problem.

\subsection{Synthetic Prior and Training Data}
\label{sec:synthetic-prior-training-data}

To generalize across multi-objective problems, the hypernetwork should be exposed to a large corpus of offline observations spanning diverse multi-objective problems during pretraining. However, obtaining observations at this scale from real-world systems is often impractical due to high evaluation costs, potential safety risks, privacy or proprietary restrictions. We therefore adopt prior-data pretraining \citep{nguyen2023expt,hollmann2025accurate,zhang2026tamo}, synthesizing a large and diverse collection of training data from a designed generative prior that varies the decision and objective dimensionalities, context size, and objective-landscape geometry, thereby exposing the hypernetwork to a broad range of problem structures that may arise in practice.

\subsubsection{Synthetic Prior}
\label{sec:synthetic-prior}

We adopt a Gaussian process (GP) prior to synthesize objective functions because it provides a cost-effective and flexible mechanism for generating a large collection of diverse problem landscapes. Moreover, the kernel family and its hyperparameters provide direct control over the distribution of the sampled objective functions. For every objective, we independently draw a latent function from the GP prior,
\begin{equation}
f_j \sim \mathcal{GP}(0,k_j),
\qquad
k_j \sim \operatorname{Uniform}\!\left(\mathcal{K}\right),
\qquad
j=1,\ldots,m,
\label{eq:synthetic-objective-gp}
\end{equation}
where $\mathcal{K}$ contains the RBF, Matérn-$3/2$, and Matérn-$5/2$ kernels. The RBF kernel is a universal approximator \citep{micchelli2006universal} to allow the prior to encompass a broad range of unknown objective functions, whereas the Matérn kernels complement it with rougher sample paths and different degrees of regularity. In addition, the automatic relevance determination (ARD) length scales are sampled independently to induce variation in characteristic scales and anisotropy. Together, these choices enable the synthetic prior to generate objectives with diverse landscape properties.

\subsubsection{Training Data Generation}
\label{sec:training-data-generation}

At each pretraining step, we construct a synthetic training episode in three steps. \textbf{(i) Prior configuration.} We sample the decision dimensionality $d$, the number of objectives $m$, the context size $N_c$, as well as the kernel choice and ARD length scales associated with each objective. \textbf{(ii) Function realization.} Given this configuration, we instantiate a multi-objective function $\boldsymbol{f}=(f_1,\ldots,f_m)$ from the GP prior. Each objective is represented as a reusable approximate GP path using random Fourier features (RFFs), allowing the same function realization to be evaluated consistently at the predicted solutions when computing the pretraining loss. \textbf{(iii) Context and query construction.} We sample decisions $\{\boldsymbol{x}_i\}_{i=1}^{N_c}$ uniformly from the decision space $\mathcal{X}=[0,1]^d$ and collect their observations to form the offline context $\mathcal{D}=\{(\boldsymbol{x}_i,\boldsymbol{y}_i)\}_{i=1}^{N_c}$. We further sample a set of query preferences $\boldsymbol{\Lambda}=\{\boldsymbol{\lambda}_q\}_{q=1}^{Q}$ from the simplex $\Delta^{m-1}=\{\boldsymbol{\lambda}\in\mathbb{R}_{+}^{m}\mid\|\boldsymbol{\lambda}\|_1=1\}$ to evaluate the generated Pareto solution model during pretraining. Collectively, $\boldsymbol{f}$, $\mathcal{D}$, and $\boldsymbol{\Lambda}$ define a complete synthetic training episode.

\subsection{Model Architecture}
\label{sec:model-architecture}

The hypernetwork $H_{\boldsymbol{\theta}}$ transforms a variable-sized offline context $\mathcal{D}$ into a problem-specific Pareto solution model through three modules: a fixed-width context embedding, a permutation-equivariant Transformer encoder, and a low-rank parameter decoder. Together, these modules encode the offline observations into a problem-level representation and produce instance-specific parameters, which instantiate a lightweight MLP to generate a preference-conditioned solution distribution.

\subsubsection{Context Embedding}
\label{sec:context-embedding}

Given the offline context $\mathcal{D}=\{(\boldsymbol{x}_i,\boldsymbol{y}_i)\}_{i=1}^{N_c}$, directly embedding each objective observation $\boldsymbol{y}_i$ would not explicitly characterize its quality under different trade-offs. We therefore construct a fixed-width representation using a bank of reference preferences $\{\overline{\boldsymbol{\lambda}}_j\}_{j=1}^{J}\subseteq\Delta^{m-1}$ that covers both the vertices and the interior of the preference simplex. For each observation $(\boldsymbol{x}_i,\boldsymbol{y}_i)$, its Tchebycheff scalarization signature is $\boldsymbol{g}_i=(g_{i1},\ldots,g_{iJ})$, where $g_{ij}=\max_{\ell\in\{1,\ldots,m\}}\overline{\lambda}_{j\ell}y_{i\ell}$. The resulting signature maps observations of different objective dimensionalities into a common space while exposing their quality across a shared collection of trade-offs. To accommodate different decision dimensionalities, each decision is zero-padded to $d_{\max}$ dimensions. Separate linear encoders then project the padded decision and its scalarization signature into a shared latent space, and their sum forms the corresponding context token:
\begin{equation}
\boldsymbol{e}_i
=
E_x\!\left(\operatorname{pad}(\boldsymbol{x}_i)\right)
+
E_g\!\left(\boldsymbol{g}_i\right).
\label{eq:icpsl-context-token}
\end{equation}
This construction produces one token for each offline observation rather than expanding the context over all observation--preference pairs. Consequently, the number of tokens remains $|\mathcal{D}|$, while each token summarizes the corresponding observation across the reference trade-offs.

\subsubsection{Transformer Encoder}
\label{sec:transformer-encoder}

After embedding, the context tokens are jointly processed by a Transformer encoder $T_{\boldsymbol{\tau}}$ composed of multi-head self-attention, position-wise feed-forward transformations and layer normalization. Self-attention captures dependencies among the offline observations, allowing each token to be interpreted in relation to the full context. Since $\mathcal{D}$ has no canonical ordering, no positional encoding is introduced. The encoded tokens are aggregated by average pooling to obtain a problem-level representation:
\begin{equation}
\{\boldsymbol{z}_i\}_{i=1}^{N_c}
=
T_{\boldsymbol{\tau}}\!\left(\{\boldsymbol{e}_i\}_{i=1}^{N_c}\right),
\qquad
\overline{\boldsymbol{z}}_{\mathcal{D}}
=
\frac{1}{N_c}\sum_{i=1}^{N_c}\boldsymbol{z}_i.
\label{eq:icpsl-problem-representation}
\end{equation}
Together, position-free encoding and symmetric pooling make $\overline{\boldsymbol{z}}_{\mathcal{D}}$ invariant to the ordering of the offline observations.

\subsubsection{Low-Rank Parameter Decoder}
\label{sec:low-rank-parameter-decoder}

Given the pooled problem representation $\overline{\boldsymbol{z}}_{\mathcal{D}}$, a feed-forward parameter decoder $D_{\boldsymbol{\eta}}$ produces the instance-dependent parameters $\boldsymbol{\phi}_{\mathcal{D}}$ used to instantiate a problem-specific two-hidden-layer MLP:
\begin{equation}
\begin{aligned}
\boldsymbol{\phi}_{\mathcal{D}}
&=
D_{\boldsymbol{\eta}}
\!\left(\overline{\boldsymbol{z}}_{\mathcal{D}}\right)\\
&=
\bigl\{
A_1(\mathcal{D}),A_2(\mathcal{D}),W_o(\mathcal{D}),
\boldsymbol{b}_1(\mathcal{D}),\boldsymbol{b}_2(\mathcal{D}),
\boldsymbol{b}_o(\mathcal{D})
\bigr\}.
\end{aligned}
\label{eq:icpsl-parameter-decoder}
\end{equation}
Generating the hidden-layer weight matrices densely would make the decoder output unnecessarily large. Instead, the decoder predicts the problem-dependent low-rank factors $A_1(\mathcal{D})$ and $A_2(\mathcal{D})$, which are combined with the globally learned projections $S_1$ and $S_2$ to construct the hidden-layer weight matrices $W_1(\mathcal{D})$ and $W_2(\mathcal{D})$:
\begin{equation}
W_1(\mathcal{D})=A_1(\mathcal{D})S_1,
\qquad
W_2(\mathcal{D})=A_2(\mathcal{D})S_2.
\label{eq:icpsl-low-rank-weights}
\end{equation}
The factors $A_1(\mathcal D)$ and $A_2(\mathcal D)$, the full output matrix $W_o(\mathcal D)$, and all biases are generated directly from the current context. In contrast, $S_1$ and $S_2$ are global parameters learned during pretraining and shared across problem instances.

\subsubsection{Solution Model Instantiation}
\label{sec:solution-model-instantiation}

To accommodate potential discontinuities in the optimal preference-to-solution mapping $\boldsymbol{\lambda}\mapsto\boldsymbol{x}^{\star}(\boldsymbol{\lambda})$, ICPSL models a preference-conditioned solution distribution $q_{\boldsymbol{\phi}_{\mathcal D}}(\boldsymbol{x}\mid\boldsymbol{\lambda})$. We represent this distribution using a family of latent-conditioned component $q_{\boldsymbol{\phi}_{\mathcal D}}(\boldsymbol{x}\mid\boldsymbol{\lambda},z)$ indexed by $z\sim U([0,1])$, allowing probability mass to be assigned to multiple high-quality solution regions associated with the same preference. To parameterize each component, the $k$-th realization $z_k$ of the latent variable $z$ is concatenated with a preference zero-padded to $m_{\max}$ dimensions, yielding $\boldsymbol{v}_k(\boldsymbol{\lambda})=[\operatorname{pad}(\boldsymbol{\lambda});z_k]$. The decoded parameters $\boldsymbol{\phi}_{\mathcal D}$ instantiate a problem-specific MLP $h_{\boldsymbol{\phi}_{\mathcal D}}$ that processes this input as
\begin{align}
\boldsymbol{u}_{1k}
&=
\rho\!\left(
\boldsymbol{v}_k(\boldsymbol{\lambda})W_1(\mathcal{D})
+\boldsymbol{b}_1(\mathcal{D})
\right),
\label{eq:icpsl-generated-layer-one}\\
\boldsymbol{u}_{2k}
&=
\rho\!\left(
\boldsymbol{u}_{1k}W_2(\mathcal{D})
+\boldsymbol{b}_2(\mathcal{D})
\right),
\label{eq:icpsl-generated-layer-two}\\
\left[\boldsymbol{\mu}_k,\boldsymbol{\ell}_k\right]
&=
\boldsymbol{u}_{2k}W_o(\mathcal{D})
+\boldsymbol{b}_o(\mathcal{D}),
\label{eq:icpsl-generated-output}
\end{align}
where $\rho$ denotes ReLU. The outputs $\boldsymbol{\mu}_k$ and $\boldsymbol{\ell}_k$ parameterize the Gaussian mean and lower-triangular scale matrix, respectively. After converting $\boldsymbol{\ell}_k$ into a valid Cholesky factor $L_k$, these outputs define the latent-conditioned component
\begin{equation}
q_{\boldsymbol{\phi}_{\mathcal D}}
\!\left(\boldsymbol{x}\mid\boldsymbol{\lambda},z_k\right)
=
\operatorname{LogitNormal}
\!\left(\boldsymbol{x};\boldsymbol{\mu}_k,L_kL_k^\top\right).
\label{eq:icpsl-latent-conditioned-component}
\end{equation}

Finally, the preference-conditioned solution distribution is obtained by marginalizing these components over $z$. In practice, we approximate this integral using $K$ randomized Sobol points $\{z_k\}_{k=1}^{K}$:
\begin{equation}
q_{\boldsymbol{\phi}_{\mathcal{D}}}
\!\left(\boldsymbol{x}\mid\boldsymbol{\lambda}\right)
=
\int q_{\boldsymbol{\phi}_{\mathcal{D}}}
\!\left(\boldsymbol{x}\mid\boldsymbol{\lambda},z\right)p(z)\,\mathrm{d}z
\approx
\frac{1}{K}\sum_{k=1}^{K}
q_{\boldsymbol{\phi}_{\mathcal{D}}}
\!\left(\boldsymbol{x}\mid\boldsymbol{\lambda},z_k\right).
\label{eq:icpsl-conditional-mixture}
\end{equation}
The resulting distribution can allocate probability mass to multiple disconnected low-energy regions, thereby accommodating discontinuities in the optimal preference-to-solution mapping that may be difficult for a deterministic continuous solution model to represent. Appendix~\ref{sec:preference-conditioned-solution-distribution-modeling} further details the construction of the latent-conditioned components and their marginalization.

\subsection{Objective-Based Pretraining}
\label{sec:objective-based-pretraining}

ICPSL is pretrained directly using evaluations of the sampled synthetic objective functions, without requiring precomputed Pareto-optimal solutions as supervision. For a query preference $\boldsymbol{\lambda}$, we define the Tchebycheff energy in the normalized objective space
\begin{equation}
s_{\boldsymbol{\lambda}}(\boldsymbol{x})
=
\max_{j\in\{1,\ldots,m\}}
\lambda_j f_j(\boldsymbol{x}).
\label{eq:icpsl-tchebycheff-energy}
\end{equation}
Ideally, the preference-conditioned solution distribution should concentrate its probability mass on high-quality solutions under the queried preference. We therefore define the target preference-conditioned solution distribution as the Gibbs distribution induced by this energy,
\begin{equation}
p_{\tau}^{\star}
\!\left(\boldsymbol{x}\mid\boldsymbol{\lambda},\boldsymbol{f}\right)
=
\frac{\exp\!\left(-s_{\boldsymbol{\lambda}}(\boldsymbol{x})/\tau\right)}
{Z(\boldsymbol{\lambda},\boldsymbol{f})},
\qquad
Z(\boldsymbol{\lambda},\boldsymbol{f})
=
\int_{\mathcal{X}}
\exp\!\left(-s_{\boldsymbol{\lambda}}(\boldsymbol{u})/\tau\right)
\mathrm{d}\boldsymbol{u}.
\label{eq:icpsl-energy-distribution}
\end{equation}
The temperature $\tau>0$ controls its concentration, with smaller values placing more probability mass around low-energy solutions. Given the learned preference-conditioned solution distribution $q_{\boldsymbol{\phi}_{\mathcal{D}}}(\boldsymbol{x}\mid\boldsymbol{\lambda})$, ICPSL minimizes the free-energy loss
\begin{equation}
\mathcal{L}
=
\mathbb{E}_{\boldsymbol{x}\sim q_{\boldsymbol{\phi}_{\mathcal{D}}}(\boldsymbol{x}\mid\boldsymbol{\lambda})}
\!\left[s_{\boldsymbol{\lambda}}(\boldsymbol{x})\right]
+
\tau\,
\mathbb{E}_{\boldsymbol{x}\sim q_{\boldsymbol{\phi}_{\mathcal{D}}}(\boldsymbol{x}\mid\boldsymbol{\lambda})}
\!\left[
\log q_{\boldsymbol{\phi}_{\mathcal{D}}}
\!\left(\boldsymbol{x}\mid\boldsymbol{\lambda}\right)
\right].
\label{eq:icpsl-pretraining-loss}
\end{equation}
The first expectation shifts probability mass toward solutions with low Tchebycheff energy, whereas the second is the temperature-weighted negative differential entropy of the preference-conditioned solution distribution and discourages premature concentration. As derived in Appendix~\ref{sec:free-energy-loss-kl-interpretation}, minimizing $\mathcal{L}$ is equivalent to minimizing the KL divergence between the learned preference-conditioned solution distribution and the Tchebycheff-energy-induced target distribution.

During pretraining, we estimate both expectations using reparameterized samples, evaluate their Tchebycheff energies through the synthetic objective paths, and compute the marginal log density by aggregating each sample's density across all latent-conditioned components. This objective requires neither Pareto-optimal solution labels nor a separate optimizer for the sampled training problems. Appendix~\ref{sec:monte-carlo-free-energy-loss} provides the complete Monte Carlo estimator.

\subsection{In-Context Inference}
\label{sec:in-context-inference}

Once hypernetwork pretraining is complete, inference for a target problem operates directly on the generated problem-specific solution model $h_{\boldsymbol{\phi}_{\mathcal D}}$. For each queried preference, the model defines a preference-conditioned solution distribution from which a single predicted solution is produced for decision making. We consider three inference strategies: maximum a posteriori (ICPSL-MAP), component-averaged mean (ICPSL-CAM), and mixture mean (ICPSL-MM).

\textbf{ICPSL-MAP.} This strategy draws one reparameterized candidate $\boldsymbol{x}_k$ from every latent-conditioned component and evaluates each candidate under the full preference-conditioned solution density. It then selects
\begin{equation}
\hat{\boldsymbol{x}}_{\mathrm{MAP}}(\boldsymbol{\lambda})
=
\underset{\boldsymbol{x}\in\{\boldsymbol{x}_1,\ldots,\boldsymbol{x}_K\}}
{\arg\max}\;
q_{\boldsymbol{\phi}_{\mathcal{D}}}
\!\left(\boldsymbol{x}\mid\boldsymbol{\lambda}\right).
\label{eq:icpsl-map-inference}
\end{equation}
This sample-based approximate MAP rule favors a locally high-density solution while accounting for overlap among all mixture components.

\textbf{ICPSL-CAM.} This strategy averages the sigmoid-transformed component means. Specifically, it applies the sigmoid directly to each pre-sigmoid mean and computes
\begin{equation}
\hat{\boldsymbol{x}}_{\mathrm{CAM}}(\boldsymbol{\lambda})
=
\frac{1}{K}
\sum_{k=1}^{K}
\sigma\!\left(\boldsymbol{\mu}_k(\boldsymbol{\lambda})\right).
\label{eq:icpsl-cam-inference}
\end{equation}
This deterministic rule avoids Gaussian sampling and density evaluation, making it the least expensive way to aggregate all mixture components into one prediction.

\textbf{ICPSL-MM.} This strategy estimates the mean of the complete preference-conditioned solution distribution. It draws $R$ Gaussian samples from each latent-conditioned component and averages the corresponding bounded decisions:
\begin{equation}
\hat{\boldsymbol{x}}_{\mathrm{MM}}(\boldsymbol{\lambda})
=
\frac{1}{KR}
\sum_{k=1}^{K}
\sum_{r=1}^{R}
\sigma\!\left(
\boldsymbol{\mu}_k(\boldsymbol{\lambda})
+L_k(\boldsymbol{\lambda})\boldsymbol{\epsilon}_{kr}
\right),
\qquad
\boldsymbol{\epsilon}_{kr}\sim\mathcal{N}(\boldsymbol{0},I).
\label{eq:icpsl-mm-inference}
\end{equation}
Unlike ICPSL-CAM, this Monte Carlo estimate incorporates each component's covariance after the nonlinear sigmoid transformation. All three inference strategies operate solely on the generated solution model and require no parameter updates, preference-wise optimization, or additional evaluations of the true objectives.

%% file: sections/04_experiments.tex
\section{Experiments}
\label{sec:experiments}

\subsection{Experimental Setting}
\label{sec:experimental-setting}

\textbf{Baselines.} We compare ICPSL with representative baselines, including the implicit method ParetoFlow~\citep{yuan2025paretoflow}, the regression-based explicit method PSLR~\citep{amos2023tutorial}, and the objective-based explicit methods PSLO~\citep{lin2022pareto} and DOMOO~\citep{zhu2026domoo}. Detailed baseline configurations are provided in Appendix~\ref{sec:baseline-implementation-details}.

\textbf{Benchmarks.} We evaluate all methods on a diverse suite of benchmarks covering synthetic functions ~\citep{blank2020pymoo} and real-world engineering design problems ~\citep{tanabe2020easy}. For each problem, we construct an offline context of 10 observations using Latin hypercube sampling (LHS), scrambled Sobol sampling, or uniform random sampling. We conduct 10 independent runs for each sampling method, giving all algorithms the same context within each run. Further benchmark details are provided in Appendix~\ref{sec:benchmark-implementation-details}.

\textbf{Evaluation Protocol.} We evaluate the true objectives after inference and measure solution quality by hypervolume (HV) in the normalized objective space with reference point $\boldsymbol{r}=[1.1,...,1.1]^T\in\mathbb{R}^{m}$, where higher values are better. Over $10$ independent runs, we report the mean HV and the half-width of its $95\%$ confidence interval, together with wall-clock times for problem-specific model construction and solution generation across all query preferences. Further details are provided in Appendix~\ref{sec:evaluation-protocol-details}.

\textbf{Implementation.} For ParetoFlow, PSLO, and DOMOO, we use the authors' official implementations. ICPSL and PSLR are implemented in PyTorch. Baseline code sources and further implementation details are provided in Appendix~\ref{sec:implementation-details}.

\subsection{Main Results}
\label{sec:experimental-results}

\subsubsection{Optimization Performance}
We evaluate all methods under offline contexts generated by LHS, Sobol, and uniform sampling, with lower average ranks indicating better optimization performance. As shown in Table~\ref{tab:algorithm-rank-summary}, ICPSL-MM achieves the best overall average rank and the most wins, ranking first in 15 of the 30 cases.

\Needspace{0.45\textheight}
\begin{wraptable}{r}{0.68\columnwidth}
\vspace{-\intextsep}
\centering
\caption{Average ranks, wins, and top-3 counts across all benchmark instances with LHS, Sobol, and uniform context distributions.}
\label{tab:algorithm-rank-summary}
\renewcommand{\arraystretch}{1.2}
\setlength{\tabcolsep}{3pt}
\resizebox{\linewidth}{!}{%
\begin{tabular}{ccccccc}
\toprule[1pt]
Algorithm & LHS & Sobol & Uniform & Overall & Wins & Top-3 cases \\
\midrule
ParetoFlow & 4.50 & 4.50 & 4.20 & 4.40 & 5/30 & 9/30 \\
PSLR & 3.70 & 3.40 & 3.80 & 3.63 & 2/30 & 18/30 \\
PSLO & 5.10 & 5.20 & 5.80 & 5.37 & 2/30 & 5/30 \\
DOMOO & 4.30 & \cellcolor{gray!50}\textbf{3.20} & 5.10 & 4.20 & 5/30 & 10/30 \\
\midrule
ICPSL-MAP & 5.00 & 4.80 & 4.20 & 4.67 & 1/30 & 7/30 \\
ICPSL-CAM & \cellcolor{gray!50}\textbf{2.70} & 3.70 & 2.60 & 3.00 & 0/30 & \cellcolor{gray!50}\textbf{23/30} \\
ICPSL-MM & \cellcolor{gray!50}\textbf{2.70} & \cellcolor{gray!50}\textbf{3.20} & \cellcolor{gray!50}\textbf{2.30} & \cellcolor{gray!50}\textbf{2.73} & \cellcolor{gray!50}\textbf{15/30} & 18/30 \\
\bottomrule[1pt]
\end{tabular}%
}
\end{wraptable}

For each sampling method, ICPSL-MM obtains the best average rank under the uniform context distribution and shares the best rank under both LHS and Sobol, demonstrating consistent performance across different offline observation patterns. ICPSL-CAM is also competitive, sharing the best LHS rank and achieving the highest top-3 count of 23/30. These results indicate that amortizing Pareto solution-model construction across problems does not compromise optimization quality. Complete performance results are provided in Appendix~\ref{sec:detailed-main-results}.

\WFclear

\begin{figure}[!h]
\centering
\includegraphics[width=\textwidth]{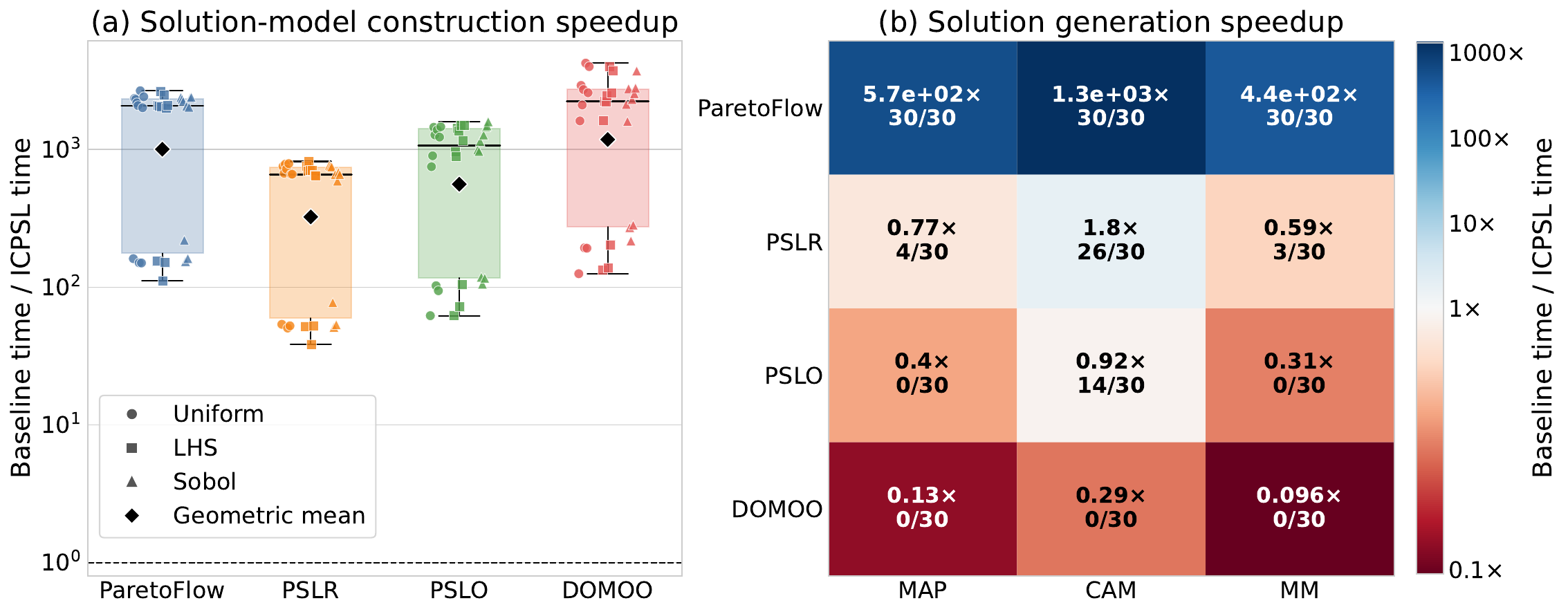}
\caption{Runtime speedups of ICPSL over the baselines for problem-specific Pareto solution-model construction and preference-conditioned solution generation.}
\label{fig:exp-runtime-speedup}
\end{figure}

\subsubsection{Computational Efficiency}
Beyond solution quality, we record the wall-clock times required for problem-specific Pareto solution-model construction and preference-conditioned solution generation. Figure~\ref{fig:exp-runtime-speedup} reports baseline-to-ICPSL runtime ratios, where values above one indicate that ICPSL is faster. For model construction, ICPSL provides orders-of-magnitude speedups over all four baselines across all 30 cases because a single hypernetwork forward pass replaces problem-specific training. At inference, ICPSL-CAM offers the best trade-off among the three variants, it achieves a averaged speedup of $1.3\times 10^3$ over ParetoFlow in 30/30 cases and $1.8\times$ over PSLR in 26/30 cases. ParetoFlow has the longest inference time because it has to perform solution search for all queried preferences. Complete timing results are provided in Appendix~\ref{sec:detailed-main-results}.

\subsection{Additional Analysis}
\label{sec:additional-analysis}

\textbf{Evolution of Solution-Model Quality.} We track solution-model quality across checkpoints saved every 30 epochs on three unseen problems during pretraining (Figure~\ref{fig:exp-hv-vs-epoch} in Appendix~\ref{sec:evolution-solution-model-quality-results}). The results show that the solution quality generally increases with pretraining, although the trajectories are not monotonic. On VLMOP2 and ABC, ICPSL-CAM and ICPSL-MM consistently outperform ICPSL-MAP, and their improvement is particularly pronounced on VLMOP2. This pattern suggests that aggregating across latent-conditioned components helps convert the learned distribution into a more effective point prediction. The less stable RE23 trajectories under uniform contexts indicate that these gains are not uniform across offline observation patterns.

\textbf{Sensitivity to Context Size.} We examine how solution quality varies with different context sizes (Figure~\ref{fig:exp-hv-vs-context-size} in Appendix~\ref{sec:sensitivity-context-size-results}). The results indicate that solution quality generally improves as the context grows, but the gains vary across benchmarks and inference strategies, with some trajectories rising gradually and others fluctuating or leveling off. This suggests that additional observations can provide useful problem-specific information, although their benefits are not uniformly concentrated at small context sizes. The smallest evaluation contexts also contain fewer observations than those used during pretraining, probing adaptation to greater data scarcity. Continued gains with larger contexts indicate that additional observations can still refine the generated model.

\textbf{Generalization to OOD Dimensions.} We evaluate generalization beyond the dimensionalities sampled during pretraining. As reported in Appendix~\ref{sec:ood-dimension-results}, ICPSL-MM achieves the highest hypervolume on all three five-dimensional problems under every context distribution, whereas ParetoFlow is strongest on DTLZ4. This contrast suggests that transferring learned trade-off structure to an unseen number of objectives is more challenging in these tests than extending to an additional decision variable, even though the architecture accommodates both dimensions. ICPSL nonetheless retains the fastest model construction across all OOD settings, achieving a substantial computational advantage while delivering leading or competitive solution quality.

%% file: sections/05_conclusion.tex
\section{Conclusion}
\label{sec:conclusion}

We introduced ICPSL, which pretrains a foundation-style hypernetwork on a large and diverse set of synthetic prior data to generate a reusable and problem-specific solution model from limited offline context in one forward pass, amortizing both model construction across problems and solution search across preferences. Across synthetic and engineering benchmarks, ICPSL achieves high-quality Pareto solutions while providing substantial speedups in solution-model construction.

Our current evaluation primarily covers low-dimensional problems, leaving broader dimensional settings for future study. Marginal-density evaluation also becomes more costly as the number of latent components grows. Future work could diversify synthetic pretraining problems and develop more efficient marginal-density estimators to extend ICPSL to larger-scale settings.

%% file: sections/iclr_ai_use_statement.tex
\subsection*{AI use statement}


In this work, we used generative AI tools for helping implementing baseline methods and interpreting experimental results. We have not used generative AI tools for generating synthetic data sets, helping develop theoretical models or conceptual frameworks, formulating mathematical claims, providing critical ingredients for proving mathematical claims, assisting in the writing of proofs, designing or providing feedback on research methodology or experiments, and proposing or refining hypotheses, assisting with translation, cleaning and reformating dataset, supporting qualitative and thematic data analysis are not applicable to this work. Additionally, we used generative AI tools for creating or modifying scientific figures, summarizing or analyzing existing literature, drafting parts of the paper, improving its readability, and suggesting its structure. We manually reviewed all AI-assisted work, checked baseline implementations against experimental outputs, and verified figures, literature summaries, and result interpretations against their underlying materials. We take responsibility for the final content of this work, including text, claims, and artifacts produced with the aid of generative AI.

%% file: sections/iclr_reproducibility_statement.tex
\subsection*{Reproducibility statement}


Appendix~\ref{sec:implementation-details} documents the implementation details used in our experiments. Table~\ref{tab:app-baseline-code-sources} identifies the baseline code sources, Appendix~\ref{sec:benchmark-implementation-details} specifies the benchmark configurations and implementations, and Appendix~\ref{sec:evaluation-protocol-details} describes the evaluation protocol.

%% file: sections/06_appendix.tex
\appendix

\section*{Appendix}
\label{sec:appendix}

\section{Additional Related Work}
\label{app:additional-related-work}

\subsection{Universal Regression Models}
Universal regression models seek to transfer function inference across tasks rather than fit a new predictor from scratch for every function. Conditional Neural Processes (CNPs) encode an unordered set of input--output observations into a representation used to predict target values \citep{pmlr-v80-garnelo18a}, while Transformer Neural Processes (TNPs) formulate uncertainty-aware meta-regression as sequence modeling \citep{pmlr-v162-nguyen22b}. Prior-Data Fitted Networks (PFNs) instead pretrain a Transformer on datasets sampled from a prior to approximate Bayesian posterior prediction in context \citep{muller2022transformers}. TabPFN develops this paradigm into a foundation model for small tabular datasets \citep{hollmann2025accurate}. Subsequent work extends broadly pretrained regression to zero-shot forecasting with ForecastPFN \citep{dooley2023forecastpfn} and to heterogeneous numerical tasks represented as text with OmniPred \citep{song2024omnipred}. These models predict responses or predictive distributions, so their use in optimization generally requires a subsequent search procedure. ICPSL instead constructs a preference-conditioned solution model directly from offline observations.

\subsection{Hypernetworks}
Hypernetworks generate the parameters of another neural network, providing a mechanism for conditioning an entire model on auxiliary information \citep{ha2017hypernetworks}. SMASH conditions a hypernetwork on candidate architectures to generate their weights, thereby amortizing the cost of evaluating architectures during neural architecture search \citep{brock2018smash}. HyperFormer instead uses shared hypernetworks conditioned on the task, Transformer layer, and adapter position to generate adapter parameters for parameter-efficient multi-task adaptation \citep{karimi2021hyperformer}. Beyond conditioning on architecture or task identifiers, hypernetworks can also generate model parameters directly from observed examples. HyperTransformer produces few-shot convolutional-network weights from support examples \citep{pmlr-v162-zhmoginov22a}, while MotherNet combines prior-data pretraining with a Transformer hypernetwork to generate a compact classifier for a new tabular dataset in a single forward pass without dataset-specific gradient descent \citep{mueller2025mothernet}. Unlike prior hypernetwork approaches, ICPSL targets offline multi-objective optimization by generating a problem-specific solution model from offline observations to predict solutions for queried preferences. ICPSL further differs in its objective-based pretraining, which evaluates sampled solutions on synthetic objectives without requiring Pareto-optimal solution labels.

\subsection{Foundation-Style Optimization}
Several methods transfer optimization experience across problem instances through pretrained sequence models. OptFormer learns function prediction and optimization-policy behavior from heterogeneous hyperparameter-optimization trajectories \citep{chen2022optformer}. PFNs4BO performs in-context Bayesian inference from a synthetic prior for Bayesian optimization \citep{pmlr-v202-muller23a}. ExPT pretrains a few-shot experimental-design policy on synthetic functions \citep{nguyen2023expt}. End-to-end meta-BO jointly learns a Transformer Neural Process surrogate and acquisition mechanism through reinforcement learning \citep{maraval2023metabo}. This direction has also been extended to multiple objectives. BOFormer uses non-Markovian deep Q-learning and Transformer sequence modeling to learn a non-myopic acquisition policy for multi-objective Bayesian optimization \citep{hung2025boformer}. TAMO pretrains a fully amortized policy that proposes the next design from the query history in a single forward pass across varying input and objective dimensions \citep{zhang2026tamo}. These approaches primarily target sequential candidate selection or experimental design. In contrast, ICPSL addresses offline multi-objective optimization without additional objective evaluations and directly constructs a reusable solution model for arbitrary queried preferences.

\section{Details of Preference-Conditioned Solution Distribution Modeling}
\label{sec:preference-conditioned-solution-distribution-modeling}

This section provides further details on preference-conditioned solution distribution modeling in ICPSL. We first describe how the problem-specific MLP parameterizes the latent-conditioned components. We then explain how component marginalization yields the preference-conditioned solution distribution and enables it to represent multiple solution modes. Finally, we discuss the main challenges associated with marginal-density evaluation and latent collapse.

\subsection{Latent-Conditioned Component Modeling}
\label{sec:latent-conditioned-component-modeling}

Given the offline context $\mathcal{D}$ from the target problem, the hypernetwork first produces the problem-dependent parameters $\boldsymbol{\phi}_{\mathcal{D}}=H_{\boldsymbol{\theta}}(\mathcal{D})$, which instantiates the problem-specific MLP $h_{\boldsymbol{\phi}_{\mathcal{D}}}$. For a preference $\boldsymbol{\lambda}$ and a scalar latent variable $z$, this MLP outputs
\begin{equation}
h_{\boldsymbol{\phi}_{\mathcal{D}}}(\boldsymbol{\lambda},z)
=
\left(
\boldsymbol{\mu}(\boldsymbol{\lambda},z),
\widetilde{L}(\boldsymbol{\lambda},z)
\right),
\label{eq:app-generated-distribution-parameters}
\end{equation}
where $\boldsymbol{\mu}(\boldsymbol{\lambda},z)\in\mathbb{R}^{d}$ is a Gaussian mean and $\widetilde{L}(\boldsymbol{\lambda},z)\in\mathbb{R}^{d\times d}$ is a raw lower-triangular scale matrix. We obtain the numerically stable Cholesky factor $L(\boldsymbol{\lambda},z)$ by retaining the strictly lower-triangular entries of $\widetilde{L}(\boldsymbol{\lambda},z)$ and parameterizing its diagonal as
\begin{equation}
L_{rr}(\boldsymbol{\lambda},z)
=
\operatorname{softplus}
\!\left(\widetilde{L}_{rr}(\boldsymbol{\lambda},z)\right)
+\delta,
\qquad
r=1,\ldots,d,
\label{eq:app-positive-cholesky-diagonal}
\end{equation}
where $\delta>0$ is a small constant to ensure that $\Sigma(\boldsymbol{\lambda},z)=L(\boldsymbol{\lambda},z)L(\boldsymbol{\lambda},z)^{\top}$ is a valid and numerically stable covariance matrix. A reparameterized sample is generated as
\begin{equation}
\boldsymbol{a}
=
\boldsymbol{\mu}(\boldsymbol{\lambda},z)
+L(\boldsymbol{\lambda},z)\boldsymbol{\epsilon},
\qquad
\boldsymbol{\epsilon}\sim\mathcal{N}(\boldsymbol{0},I),
\qquad
\boldsymbol{x}=\sigma(\boldsymbol{a}),
\label{eq:app-logistic-normal-reparameterization}
\end{equation}
where the element-wise sigmoid maps the unconstrained Gaussian sample to the normalized decision domain $(0,1)^d$, inducing the latent-conditioned logit-normal component
\begin{equation}
q_{\boldsymbol{\phi}_{\mathcal{D}}}
\!\left(\boldsymbol{x}\mid\boldsymbol{\lambda},z\right)
=
\operatorname{LogitNormal}
\!\left(
\boldsymbol{x};
\boldsymbol{\mu}(\boldsymbol{\lambda},z),
\Sigma(\boldsymbol{\lambda},z)
\right).
\label{eq:app-conditional-logistic-normal}
\end{equation}
This distribution forms a latent-conditioned component of the preference-conditioned solution distribution. Its location and local geometry depend jointly on the queried preference and latent variable.

\subsection{Component Marginalization and Its Role}
\label{sec:component-marginalization-role}

ICPSL places a uniform prior over the latent variable $z\sim U([0,1])$, and defines its preference-conditioned solution distribution by marginalization:
\begin{equation}
q_{\boldsymbol{\phi}_{\mathcal{D}}}
\!\left(\boldsymbol{x}\mid\boldsymbol{\lambda}\right)
=
\int_{0}^{1}
q_{\boldsymbol{\phi}_{\mathcal{D}}}
\!\left(\boldsymbol{x}\mid\boldsymbol{\lambda},z\right)
\mathrm{d}z.
\label{eq:app-continuous-latent-mixture}
\end{equation}
This preference-conditioned solution distribution can be viewed as a continuous mixture of logistic-normal components. Different regions of the latent domain may correspond to distinct solution modes, and the probability mass assigned to each mode is determined by the proportion of latent values mapped to that mode. In practice, we approximate the integral using a randomized one-dimensional Sobol design $\{z_k\}_{k=1}^{K}$:
\begin{equation}
q_{\boldsymbol{\phi}_{\mathcal{D}}}
\!\left(\boldsymbol{x}\mid\boldsymbol{\lambda}\right)
\approx
\frac{1}{K}
\sum_{k=1}^{K}
q_{\boldsymbol{\phi}_{\mathcal{D}}}
\!\left(\boldsymbol{x}\mid\boldsymbol{\lambda},z_k\right).
\label{eq:app-finite-latent-mixture}
\end{equation}
For notational simplicity, we use $q$ to denote this finite-mixture approximation hereafter. The same latent design is shared across the query preferences of a problem, so each $z_k$ retains a consistent component identity as $\boldsymbol{\lambda}$ varies.

The latent variable is essential when nearby preferences correspond to well-separated low-energy solution regions. A deterministic continuous model that maps each preference directly to a single solution may interpolate between these regions, potentially yielding decisions with high Tchebycheff values. By contrast, distinct subsets of the latent domain can allocate probability mass to different regions and smoothly transfer that mass as the preference changes. ICPSL therefore models a preference-conditioned solution distribution over low-Tchebycheff solutions rather than forcing a continuous network to reproduce a single-valued optimal solution mapping $\boldsymbol{\lambda}\mapsto\boldsymbol{x}^{\star}(\boldsymbol{\lambda})$ that may be discontinuous.

\subsection{Marginalization Challenges}
\label{sec:marginalization-challenges}

The main technical challenge lies in evaluating the density of the preference-conditioned solution distribution, as the continuous marginalization in Equation~\ref{eq:app-continuous-latent-mixture} generally has no closed-form solution. ICPSL addresses this challenge using the finite mixture in Equation~\ref{eq:app-finite-latent-mixture}, which approximates the continuous marginal with an explicit average over $K$ latent-conditioned components. The marginal density of each generated sample is therefore computed by evaluating it under all $K$ components and aggregating the resulting densities. With one sample drawn from each component, this procedure requires $K^2$ pairwise density evaluations for every problem--preference pair, which are computed in chunks to control memory usage. The complete Monte Carlo computation is detailed in Section~\ref{sec:monte-carlo-free-energy-loss}.

The second challenge is latent collapse when a preference admits multiple separated low-energy solution regions. If different values of $z$ produce nearly identical distribution parameters, the resulting components may concentrate on the same region rather than represent distinct modes. However, when the optimal solution mapping is unique and continuous, component overlap is natural because there are no distinct solution modes to represent. The marginal-entropy regularization described in Appendix~\ref{sec:tchebycheff-free-energy-loss-details} mitigates harmful collapse by discouraging premature concentration of the preference-conditioned solution distribution.

\section{Details of the Tchebycheff Free-Energy Loss}
\label{sec:tchebycheff-free-energy-loss-details}

This section derives the free-energy loss based on Tchebycheff aggregation for training the preference-conditioned solution distribution model and describes its Monte Carlo implementation.

\subsection{Free-Energy Loss and KL Interpretation}
\label{sec:free-energy-loss-kl-interpretation}

For a preference $\boldsymbol{\lambda}\in\Delta^{m-1}$ and a synthetic multi-objective function $\boldsymbol{f}$, ICPSL aggregates the objectives into the hard Tchebycheff energy
\begin{equation}
s_{\boldsymbol{\lambda}}(\boldsymbol{x})
=
\max_{j\in\{1,\ldots,m\}}
\lambda_j f_j(\boldsymbol{x}).
\label{eq:app-tchebycheff-energy}
\end{equation}
Ideally, a preference-conditioned solution distribution should concentrate its probability mass on high-quality solutions under the queried preference. We therefore define the target  preference-conditioned solution distribution as the Gibbs distribution induced by this energy:
\begin{equation}
p_{\tau}^{\star}
\!\left(\boldsymbol{x}\mid\boldsymbol{\lambda},\boldsymbol{f}\right)
=
\frac{
\exp\!\left(-s_{\boldsymbol{\lambda}}(\boldsymbol{x})/\tau\right)
}{
Z(\boldsymbol{\lambda},\boldsymbol{f})
},
\qquad
Z(\boldsymbol{\lambda},\boldsymbol{f})
=
\int_{\mathcal{X}}
\exp\!\left(-s_{\boldsymbol{\lambda}}(\boldsymbol{u})/\tau\right)
\mathrm{d}\boldsymbol{u}.
\label{eq:app-energy-target-density}
\end{equation}
The temperature $\tau>0$ controls the concentration of this distribution. As $\tau$ decreases, probability mass concentrates around the global minimizers of the Tchebycheff energy, while larger values preserve broader coverage of near-optimal regions.

Given the learned preference-conditioned solution distribution $q_{\boldsymbol{\phi}_{\mathcal{D}}}(\boldsymbol{x}\mid\boldsymbol{\lambda})$, ICPSL minimizes the loss
\begin{equation}
\mathcal{L}
=
\mathbb{E}_{\boldsymbol{x}\sim q_{\boldsymbol{\phi}_{\mathcal{D}}}(\boldsymbol{x}\mid\boldsymbol{\lambda})}
\!\left[s_{\boldsymbol{\lambda}}(\boldsymbol{x})\right]
+
\tau\,
\mathbb{E}_{\boldsymbol{x}\sim q_{\boldsymbol{\phi}_{\mathcal{D}}}(\boldsymbol{x}\mid\boldsymbol{\lambda})}
\!\left[
\log q_{\boldsymbol{\phi}_{\mathcal{D}}}
\!\left(\boldsymbol{x}\mid\boldsymbol{\lambda}\right)
\right].
\label{eq:app-energy-loss}
\end{equation}
To characterize how this loss aligns the learned preference-conditioned solution distribution with the target distribution, we express their discrepancy using the KL divergence. Starting from its definition and substituting the target density in Equation~\ref{eq:app-energy-target-density}, we obtain
\begin{equation}
\begin{aligned}
&\mathrm{KL}\!\left(
q_{\boldsymbol{\phi}_{\mathcal{D}}}(\cdot\mid\boldsymbol{\lambda})
\,\|\,
p_{\tau}^{\star}(\cdot\mid\boldsymbol{\lambda},\boldsymbol{f})
\right)\\
&=
\mathbb{E}_{\boldsymbol{x}\sim q_{\boldsymbol{\phi}_{\mathcal{D}}}(\boldsymbol{x}\mid\boldsymbol{\lambda})}
\!\left[
\log q_{\boldsymbol{\phi}_{\mathcal{D}}}
\!\left(\boldsymbol{x}\mid\boldsymbol{\lambda}\right)
-
\log p_{\tau}^{\star}
\!\left(\boldsymbol{x}\mid\boldsymbol{\lambda},\boldsymbol{f}\right)
\right]\\
&=
\mathbb{E}_{\boldsymbol{x}\sim q_{\boldsymbol{\phi}_{\mathcal{D}}}(\boldsymbol{x}\mid\boldsymbol{\lambda})}
\!\left[
\log q_{\boldsymbol{\phi}_{\mathcal{D}}}
\!\left(\boldsymbol{x}\mid\boldsymbol{\lambda}\right)
+\frac{s_{\boldsymbol{\lambda}}(\boldsymbol{x})}{\tau}
+\log Z(\boldsymbol{\lambda},\boldsymbol{f})
\right]\\
&=
\mathbb{E}_{\boldsymbol{x}\sim q_{\boldsymbol{\phi}_{\mathcal{D}}}(\boldsymbol{x}\mid\boldsymbol{\lambda})}
\!\left[
\log q_{\boldsymbol{\phi}_{\mathcal{D}}}
\!\left(\boldsymbol{x}\mid\boldsymbol{\lambda}\right)
\right]\\
&\quad+
\frac{1}{\tau}
\mathbb{E}_{\boldsymbol{x}\sim q_{\boldsymbol{\phi}_{\mathcal{D}}}(\boldsymbol{x}\mid\boldsymbol{\lambda})}
\!\left[s_{\boldsymbol{\lambda}}(\boldsymbol{x})\right]
+\log Z(\boldsymbol{\lambda},\boldsymbol{f}).
\end{aligned}
\label{eq:app-kl-expansion}
\end{equation}
Multiplying Equation~\ref{eq:app-kl-expansion} by $\tau$ and using Equation~\ref{eq:app-energy-loss}, we have
\begin{equation}
\begin{gathered}
\tau\,\mathrm{KL}\!\left(
q_{\boldsymbol{\phi}_{\mathcal{D}}}(\cdot\mid\boldsymbol{\lambda})
\,\|\,
p_{\tau}^{\star}(\cdot\mid\boldsymbol{\lambda},\boldsymbol{f})
\right)
=
\mathcal{L}
+\tau\log Z(\boldsymbol{\lambda},\boldsymbol{f}),\\[2pt]
\mathcal{L}
=
\tau\,\mathrm{KL}\!\left(
q_{\boldsymbol{\phi}_{\mathcal{D}}}(\cdot\mid\boldsymbol{\lambda})
\,\|\,
p_{\tau}^{\star}(\cdot\mid\boldsymbol{\lambda},\boldsymbol{f})
\right)
-\tau\log Z(\boldsymbol{\lambda},\boldsymbol{f}).
\end{gathered}
\label{eq:app-energy-loss-kl-equivalence}
\end{equation}
Because $\tau>0$ and the partition function is independent of the model parameters, Equation~\ref{eq:app-energy-loss-kl-equivalence} shows that minimizing $\mathcal{L}$ is exactly equivalent to minimizing the KL divergence between the learned preference-conditioned solution distribution and the Tchebycheff free-energy target distribution. The first expectation in Equation~\ref{eq:app-energy-loss} shifts probability mass toward low-Tchebycheff solutions, while the second is the temperature-weighted negative differential entropy of the preference-conditioned solution distribution and discourages premature concentration. Section~\ref{sec:monte-carlo-free-energy-loss} details the corresponding Monte Carlo implementation.

\subsection{Monte Carlo Estimation of the Free-Energy Loss}
\label{sec:monte-carlo-free-energy-loss}

Consider a minibatch containing problem instances indexed by $b$, query preferences indexed by $q$, latent-conditioned components indexed by $k$, and Gaussian samples within each latent-conditioned component indexed by $r$. We use $j$ to index the latent-conditioned component under which a generated sample is evaluated. For each synthetic problem, ICPSL generates $K$ randomized Sobol latent points $\{z_{bk}\}_{k=1}^{K}$ and shares the same set across its query preferences. The generated MLP produces
\begin{equation}
\left(
\boldsymbol{\mu}_{bqk},L_{bqk}
\right)
=
h_{\boldsymbol{\phi}_{\mathcal{D}_b}}
\!\left(\boldsymbol{\lambda}_{bq},z_{bk}\right),
\qquad
\boldsymbol{x}_{bqkr}
=
\sigma\!\left(
\boldsymbol{\mu}_{bqk}
+L_{bqk}\boldsymbol{\epsilon}_{bkr}
\right),
\label{eq:app-training-reparameterization}
\end{equation}
where $\boldsymbol{\epsilon}_{bkr}\sim\mathcal{N}(\boldsymbol{0},I)$ is also shared across the preferences of a problem. The synthetic GP paths are then evaluated for all generated samples to compute their Tchebycheff energies
\begin{equation}
s_{bqkr}
=
\max_{\ell\in\{1,\ldots,m_b\}}
\lambda_{bq\ell}
f_{b\ell}\!\left(\boldsymbol{x}_{bqkr}\right).
\label{eq:app-sampled-energy}
\end{equation}
The reparameterization in Equation~\ref{eq:app-training-reparameterization} allows the energy gradient to propagate through the synthetic objectives and generated decisions to the hypernetwork parameters.

The critical computation is the marginal density of each sample. Using the finite set of Sobol latent points, the marginal density of each sample is approximated as
\begin{equation}
q_{\boldsymbol{\phi}_{\mathcal{D}_b}}
\!\left(
\boldsymbol{x}_{bqkr}\mid\boldsymbol{\lambda}_{bq}
\right)
\approx
\frac{1}{K}
\sum_{j=1}^{K}
q_{\boldsymbol{\phi}_{\mathcal{D}_b}}
\!\left(
\boldsymbol{x}_{bqkr}\mid\boldsymbol{\lambda}_{bq},z_{bj}
\right),
\label{eq:app-training-finite-mixture}
\end{equation}
so every $\boldsymbol{x}_{bqkr}$ must be evaluated under all $K$ latent-conditioned components. For numerical stability, the aggregation is performed in the log domain:
\begin{equation}
\log
q_{\boldsymbol{\phi}_{\mathcal{D}_b}}
\!\left(
\boldsymbol{x}_{bqkr}\mid\boldsymbol{\lambda}_{bq}
\right)
=
\operatorname{logsumexp}_{j}
\left[
\log q_{\boldsymbol{\phi}_{\mathcal{D}_b}}
\!\left(
\boldsymbol{x}_{bqkr}
\mid\boldsymbol{\lambda}_{bq},z_{bj}
\right)
\right]
-\log K.
\label{eq:app-log-marginal-density}
\end{equation}
This all-sample-to-all-component evaluation requires $O(BQK^2R)$ density computations.

Because the Gaussian is defined in the unconstrained variable $\boldsymbol{a}$ while the loss uses the bounded decision $\boldsymbol{x}=\sigma(\boldsymbol{a})$, each latent-conditioned component density includes the sigmoid change of variables:
\begin{equation}
\begin{aligned}
\log q_{\boldsymbol{\phi}_{\mathcal{D}_b}}
\!\left(
\boldsymbol{x}_{bqkr}\mid\boldsymbol{\lambda}_{bq},z_{bj}
\right)
&=
\log\mathcal{N}
\!\left(
\operatorname{logit}(\boldsymbol{x}_{bqkr});
\boldsymbol{\mu}_{bqj},
L_{bqj}L_{bqj}^{\top}
\right)\\
&\quad-
\sum_{u=1}^{d_b}
\log\!\left[x_{bqkr,u}(1-x_{bqkr,u})\right].
\end{aligned}
\label{eq:app-logistic-normal-density}
\end{equation}
The inverse-Jacobian correction, corresponding to the second term in Equation~\ref{eq:app-logistic-normal-density}, is evaluated from the pre-sigmoid sample as
$\sum_{u=1}^{d_b}\left[\operatorname{softplus}(-a_{bqkr,u})+\operatorname{softplus}(a_{bqkr,u})\right]$,
which avoids unstable evaluations of $\log 0$ near the boundary.

Finally, combining the sampled Tchebycheff energy in Equation~\ref{eq:app-sampled-energy} with the marginal log-density approximation in Equation~\ref{eq:app-log-marginal-density} yields the Monte Carlo objective
\begin{equation}
\widehat{\mathcal{L}}
=
\frac{1}{BQKR}
\sum_{b=1}^{B}
\sum_{q=1}^{Q}
\sum_{k=1}^{K}
\sum_{r=1}^{R}
\left[
s_{bqkr}
+\tau\log
q_{\boldsymbol{\phi}_{\mathcal{D}_b}}
\!\left(
\boldsymbol{x}_{bqkr}\mid\boldsymbol{\lambda}_{bq}
\right)
\right].
\label{eq:app-monte-carlo-energy-loss}
\end{equation}
The final model uses minibatches of $B=128$ problem instances, with $Q=128$ query preferences per instance, $K=128$ latent-conditioned components per preference, and $R=1$ Gaussian sample per component, while the temperature is set to $\tau=0.015$.

\section{Implementation Details}
\label{sec:implementation-details}

\subsection{Baseline Details}
\label{sec:baseline-implementation-details}

We summarize the configurations  and implementations of each baseline below. All methods receive identical offline contexts for a given benchmark, sampling scheme, and random seed, and use the same objective normalization, query preferences, and evaluation protocol. Method-specific hyperparameters follow the official implementations.

\textbf{ParetoFlow.} ParetoFlow fits a flow-matching model to the offline decisions and a separate objective predictor for each objective, and then guides the learned flow with the predictors to generate preference-conditioned solutions~\citep{yuan2025paretoflow}. The flow-matching network has a hidden dimension of $512$ and is trained for $500$ epochs with batch size $64$ and learning rate $10^{-3}$. Each objective predictor comprises two hidden layers of width $2{,}048$ and is trained for $100$ epochs with batch size $32$, an initial learning rate of $10^{-3}$, and a multiplicative decay factor of $0.98$. During solution generation, ParetoFlow integrates the guided dynamics over $500$ steps, using $5$ offspring per preference, cosine-based neighborhoods, and a guidance coefficient of $2$.

\textbf{PSLR.} PSLR selects the offline design with the lowest augmented Tchebycheff value for each preference and uses it as the corresponding amortized regression target~\citep{amos2023tutorial}. It then trains a preference-to-solution MLP by minimizing the mean-squared error between its predictions and the regression targets. The MLP comprises two hidden layers of width $256$ followed by a sigmoid output layer. Optimization is performed for $500$ updates using Adam with learning rate $10^{-3}$, $10$ randomly sampled preferences per update, and a Tchebycheff augmentation coefficient of $0.01$.

\textbf{PSLO.} PSLO adopts the objective-based loss~\citep{lin2022pareto}. It fits an independent Gaussian process to each normalized objective and trains a preference-to-solution MLP by directly minimizing the augmented Tchebycheff scalarization of the corresponding lower confidence bounds. The confidence coefficient is set to $0.1$, and the solution MLP uses the same two-hidden-layer, width-$256$ architecture and sigmoid output as PSLR. The MLP is optimized for $500$ updates using Adam with learning rate $10^{-3}$, $10$ preferences per update, and a Tchebycheff augmentation coefficient of $0.01$.

\textbf{DOMOO.} DOMOO integrates objective surrogates, energy-based modeling of out-of-distribution risk, and nested Pareto set learning technique~\citep{zhu2026domoo}. Each objective surrogate contains two hidden layers of width $2{,}048$ and is trained for $100$ epochs with batch size $256$, an initial learning rate of $10^{-3}$, and a multiplicative decay factor of $0.98$. The energy model comprises two width-$512$ hidden layers and is trained for $50$ epochs with learning rate $10^{-3}$. The two-hidden-layer, width-$256$ Pareto solution MLP is first pretrained for $200$ epochs on the nondominated context designs and subsequently refined for $300$ updates using batches of $256$ preferences. The solution model is optimized with Adam at a learning rate of $10^{-3}$, with preferences sampled afresh from a Dirichlet distribution during the first $200$ refinement updates.

\begin{table}[!ht]
\centering
\footnotesize
\caption{Code sources used to implement the baseline methods.}
\label{tab:app-baseline-code-sources}
\renewcommand{\arraystretch}{1.15}
\begin{tabular}{@{}p{0.16\textwidth}p{0.31\textwidth}p{0.45\textwidth}@{}}
\toprule
Method & Implementation source & Code repository \\
\midrule
ParetoFlow & Official ParetoFlow implementation & \url{https://github.com/mila-iqia/ParetoFlow} \\
PSLR & Implementation in PyTorch & \url{https://github.com/pytorch/pytorch} \\
PSLO & Official PSL-MOBO implementation & \url{https://github.com/Xi-L/PSL-MOBO} \\
DOMOO & Official DOMOO implementation & \url{https://github.com/YaolinWen/DOMOO} \\
\bottomrule
\end{tabular}
\end{table}

\subsection{ICPSL Implementation Details}
\label{sec:icpsl-implementation-details}

We detail the implementation of ICPSL, covering synthetic data generation, the hypernetwork architecture, objective-based pretraining, and in-context inference. Table~\ref{tab:app-implementation-config} summarizes the corresponding numerical settings.

\begin{table}[!ht]
\centering
\caption{Key configurations of the reported ICPSL model.}
\label{tab:app-implementation-config}
\renewcommand{\arraystretch}{1.2}
\scalebox{1.1}{%
\begin{tabular}{@{}lll@{}}
\toprule[1pt]
Component & Configuration & Setting \\
\hline
\multirow{9}{*}{\shortstack[l]{Synthetic data}}
& Decision dimension & $d\sim\mathcal U\{1,\ldots,4\}$ \\
& Objectives dimension & $m\sim\mathcal U\{2,3\}$ \\
& Context size & $N_c\sim\mathcal U\{8,\ldots,64\}$ \\
& Query preferences & $Q=128$ \\
& Sobol latent points & $K=128$ \\
& Kernel family & RBF, Mat\'ern-$3/2$, or Mat\'ern-$5/2$ \\
& ARD length scale & Log-uniform on $[0.1,0.5]$ \\
& Random Fourier features & $2{,}048$ \\
& Objective sigmoid scale & $1.5$ \\
\hline
\multirow{14}{*}{Architecture}
& Maximum decision dimension & $d_{\max}=5$ \\
& Maximum objective dimension & $m_{\max}=4$ \\
& Context reference preferences & $J=128$ \\
& Transformer layers & $12$ \\
& Embedding dimension & $512$ \\
& Attention heads & $4$ \\
& Feed-forward dimension & $1{,}024$ \\
& Context pooling & Average \\
& Parameter decoder hidden layers & One layer of width $2{,}048$ \\
& Parameter decoder activation & GELU \\
& Low-rank dimension & $32$ \\
& Solution MLP hidden layers & Two layers of width $256$ \\
& Solution MLP activation & ReLU \\
& Cholesky diagonal floor & $\delta=10^{-3}$ \\
\hline
\multirow{9}{*}{Pretraining}
& Optimizer & AdamW \\
& Learning rate & $3\times10^{-5}$ \\
& Training epochs & $300$ \\
& Steps per epoch & $256$ \\
& Warmup epochs & $5$ \\
& Learning-rate schedule & Cosine decay \\
& Effective batch size & $128$ \\
& Temperature & $\tau=0.015$ \\
& Samples per component & $R=1$ \\
\bottomrule[1pt]
\end{tabular}%
}
\end{table}

\subsubsection{Synthetic Prior and Data Generation}
For every optimization step, the decision dimension, number of objectives, and GP configurations are resampled before independent objective functions are instantiated. Each GP path uses one of the three kernel families in Table~\ref{tab:app-implementation-config}, independently sampled coordinate-wise ARD length scales, and $2{,}048$ random Fourier features. Applying a sigmoid maps the sampled objectives to the unit interval while preserving repeatable evaluations of each realization.

Each synthetic problem is evaluated with the context size drawn uniformly between $8$ and $64$. Pretraining uses $128$ query preferences drawn from the simplex, and $128$ randomized one-dimensional Sobol latent values shared across those preferences. Entirely new objective functions, observations, preferences, and latent values are generated online for every training step rather than retrieved from a fixed dataset.

\subsubsection{Hypernetwork Architecture}
The context encoder uses $128$ fixed reference preferences, consisting of the simplex vertices and reproducible interior points, to construct a fixed-width Tchebycheff scalarization representation for each observation. Decisions are padded to $d_{\max}=5$ and embedded together with the scalarization signatures in a $512$-dimensional token space. Twelve Transformer layers with four attention heads and width-$1{,}024$ feed-forward blocks process the tokens without positional encoding. Average pooling then produces the permutation-invariant problem representation.

The pooled representation is passed through a parameter decoder with a single hidden layer of width $2{,}048$ and GELU activation. This decoder generates the problem-dependent low-rank factors, output matrix, and biases of the solution MLP, while rank-$32$ projections are shared across problems. The instantiated MLP has two hidden layers of width $256$ with ReLU activations and maps each padded preference--latent pair to a Gaussian mean and a lower-triangular scale factor. A softplus transformation with a floor of $10^{-3}$ ensures positive Cholesky diagonals, and only coordinates associated with the active decision dimensions are retained.

\subsubsection{Objective-Based Pretraining}
ICPSL is pretrained using the Tchebycheff free-energy objective described in Section~\ref{sec:objective-based-pretraining}. At each optimization step, the model processes an effective batch of $128$ independently sampled synthetic problems. For every problem and query preference, one reparameterized sample is drawn from each of the $128$ latent-conditioned components, evaluated on the corresponding synthetic objectives, and scored using temperature $0.015$.

We pretrain the hypernetwork for $300$ epochs, with $256$ updates per epoch, using AdamW with zero weight decay and an initial learning rate of $3\times10^{-5}$. The learning rate is warmed up linearly over the first five epochs and follows a cosine schedule thereafter. This pretraining jointly updates the context embeddings, Transformer encoder, parameter decoder, and shared low-rank projections. The complete pretraining process took approximately $21$ hours on four NVIDIA L20 GPUs.

\subsubsection{In-Context Inference}
At inference stage, the frozen hypernetwork generates the parameters of a problem-specific solution MLP model from the normalized offline context in a single forward pass. We use $128$ randomized Sobol latent values shared across query preferences. ICPSL-MAP draws one sample from each latent-conditioned component and selects the candidate with the highest marginal density. ICPSL-CAM averages the sigmoid-transformed component centers, whereas ICPSL-MM estimates the mixture mean using $64$ samples per component. The same generated MLP is used for every query preference without parameter updates and preference-wise optimization.

\subsection{Benchmark Details}
\label{sec:benchmark-implementation-details}

Our main benchmark suite comprises seven continuous synthetic functions implemented in Pymoo~\citep{blank2020pymoo}, and three engineering design problems implemented in RE test suit~\citep{tanabe2020easy}. Their detailed configurations are summarized in Table~\ref{tab:app-benchmark-config}.

\begin{table}[!ht]
\centering
\caption{Benchmark configurations for main experiments.}
\label{tab:app-benchmark-config}
\renewcommand{\arraystretch}{1.2}
\scalebox{1.1}{%
\begin{tabular}{@{}lllcc@{}}
\toprule[1pt]
Benchmark & Category & Objective construction or design task & $d$ & $m$ \\
\hline
VLMOP1 & Classical synthetic & VLMOP test function & $1$ & $2$ \\
VLMOP2 & Classical synthetic & VLMOP test function & $3$ & $2$ \\
VLMOP3 & Classical synthetic & VLMOP test function & $2$ & $3$ \\
\hline
AR & Composite synthetic & Ackley and Rosenbrock & $2$ & $2$ \\
DR & Composite synthetic & Dixon--Price and Rastrigin & $2$ & $2$ \\
ARA & Composite synthetic & Ackley and Rastrigin & $2$ & $2$ \\
ABC & Composite synthetic & Ackley, Branin, and Currin & $2$ & $3$ \\
\hline
RE23 & Engineering design & Pressure-vessel design & $4$ & $2$ \\
RE24 & Engineering design & Hatch-cover design & $2$ & $2$ \\
RE32 & Engineering design & Welded-beam design & $4$ & $3$ \\
\bottomrule[1pt]
\end{tabular}%
}
\end{table}

\Needspace{0.35\textheight}
\subsection{Evaluation Protocol Details}
\label{sec:evaluation-protocol-details}

All algorithms operate in the normalized decision domain $[0,1]^d$. Before evaluating the objectives, each benchmark maps the proposed decisions to its original domain. Each run uses $10$ offline observations generated by LHS, scrambled Sobol, or uniform sampling, and all methods receive the same context for each of the $10$ random seeds. For bi-objective problems, query preferences partition the unit interval into $200$ equal segments and include both endpoints, yielding $201$ preferences. For problems with $m>2$ objectives, the fixed design also contains $201$ preferences, including all $m$ simplex vertices and $201-m$ interior preferences sampled uniformly from the preference simplex. All methods use the same preference design for a given problem.

To determine objective-wise normalization ranges, we evaluate the objectives at a large set of sampled designs, using $10{,}000$ points from a fixed unscrambled Sobol sequence. After inference, the true objectives are evaluated at the predicted solutions to form the objective set $\mathcal Y$. We measure solution quality by hypervolume (HV) in the normalized objective space,
\begin{equation}
\operatorname{HV}(\mathcal Y;\boldsymbol r)
=
\operatorname{vol}\!\left(
\bigcup_{\boldsymbol y\in\mathcal Y}
[\boldsymbol y,\boldsymbol r]
\right),
\label{eq:experimental-hypervolume}
\end{equation}
where $\boldsymbol{r}=[1.1,...,1.1]^T\in\mathbb{R}^{m}$ is the reference point and higher values indicate better Pareto-front approximation. Over $10$ independent runs, we report mean HV with the half-width of its $95\%$ confidence interval. We also record the wall-clock times for problem-specific model construction and solution generation across all query preferences, excluding the time spent on context sampling, objective evaluation, and model loading.

\clearpage

\section{Detailed Results for Main Experiments}
\label{sec:detailed-main-results}

We report the complete per-benchmark results for solution quality and computational efficiency under LHS, Sobol, and uniform context distributions. Overall, ICPSL achieves competitive hypervolume across the benchmark suite while requiring substantially less problem-specific training time than the baselines.

\subsection{Solution Quality}

\begin{table}[!ht]
\centering
\caption{Mean hypervolume ($\uparrow$) with 95\% confidence intervals using LHS context over 10 independent runs.}
\label{tab:app-hv-lhs}
\renewcommand{\arraystretch}{1.5}
\resizebox{\textwidth}{!}{%
\begin{tabular}{lcccccccccc}
\toprule[1pt]
Algorithm & VLMOP1 & VLMOP2 & VLMOP3 & AR & DR & ARA & ABC & RE23 & RE24 & RE32 \\
\midrule
ParetoFlow & 1.195e+00(1.2e-03) & 1.412e-01(2.6e-02) & 9.452e-01(1.9e-01) & 6.669e-01(1.2e-01) & 1.084e+00(5.7e-02) & 5.181e-01(1.2e-01) & 1.854e-01(2.8e-02) & \cellcolor{gray!50}\textbf{1.190e+00(1.0e-02)} & \cellcolor{gray!50}\textbf{1.179e+00(2.0e-02)} & 1.316e+00(1.1e-02) \\
PSLR & 1.194e+00(2.6e-03) & 2.015e-01(4.3e-02) & \cellcolor{gray!50}\textbf{1.134e+00(1.5e-01)} & 6.608e-01(1.3e-01) & 1.117e+00(5.5e-02) & 5.560e-01(1.7e-01) & 2.925e-01(2.9e-02) & 1.182e+00(7.7e-03) & 1.155e+00(1.9e-02) & 1.302e+00(1.4e-02) \\
PSLO & \cellcolor{gray!50}\textbf{1.199e+00(1.1e-06)} & 1.040e-01(4.8e-02) & 9.881e-01(2.1e-01) & 7.005e-01(1.3e-01) & 1.090e+00(5.3e-02) & 5.328e-01(1.9e-01) & 2.624e-01(5.7e-02) & 8.673e-01(3.1e-01) & 9.071e-01(2.5e-01) & 1.160e+00(2.4e-01) \\
DOMOO & \cellcolor{gray!50}\textbf{1.199e+00(9.8e-06)} & 1.571e-01(6.7e-02) & 9.928e-01(2.3e-01) & 6.813e-01(1.4e-01) & 9.841e-01(1.3e-01) & 4.631e-01(1.2e-01) & 2.803e-01(3.9e-02) & 1.072e+00(2.0e-01) & 1.161e+00(5.5e-02) & \cellcolor{gray!50}\textbf{1.317e+00(2.0e-02)} \\
\midrule
ICPSL-MAP & 1.197e+00(1.5e-03) & 1.185e-01(3.4e-02) & 9.683e-01(2.0e-01) & 6.461e-01(1.3e-01) & 1.073e+00(6.4e-02) & 5.426e-01(1.3e-01) & 2.306e-01(4.6e-02) & 1.175e+00(1.2e-02) & 1.166e+00(2.2e-02) & 1.286e+00(2.0e-02) \\
ICPSL-CAM & 1.197e+00(1.2e-03) & 3.835e-01(5.0e-02) & 1.054e+00(1.7e-01) & 7.706e-01(1.2e-01) & 1.135e+00(5.3e-02) & 6.371e-01(1.3e-01) & 3.012e-01(2.3e-02) & 1.176e+00(1.1e-02) & 1.162e+00(1.5e-02) & 1.281e+00(1.9e-02) \\
ICPSL-MM & 1.197e+00(1.2e-03) & \cellcolor{gray!50}\textbf{4.087e-01(4.9e-02)} & 1.078e+00(1.4e-01) & \cellcolor{gray!50}\textbf{7.745e-01(1.2e-01)} & \cellcolor{gray!50}\textbf{1.137e+00(5.8e-02)} & \cellcolor{gray!50}\textbf{6.601e-01(1.4e-01)} & \cellcolor{gray!50}\textbf{3.115e-01(1.8e-02)} & 1.175e+00(1.1e-02) & 1.156e+00(1.5e-02) & 1.277e+00(2.1e-02) \\
\bottomrule[1pt]
\end{tabular}%
}
\end{table}

\begin{table}[!ht]
\centering
\caption{Mean hypervolume ($\uparrow$) with 95\% confidence intervals using Sobol context over 10 independent runs.}
\label{tab:app-hv-sobol}
\renewcommand{\arraystretch}{1.5}
\resizebox{\textwidth}{!}{%
\begin{tabular}{lcccccccccc}
\toprule[1pt]
Algorithm & VLMOP1 & VLMOP2 & VLMOP3 & AR & DR & ARA & ABC & RE23 & RE24 & RE32 \\
\midrule
ParetoFlow & 1.192e+00(3.2e-03) & 1.672e-01(4.7e-02) & 9.878e-01(1.9e-01) & 5.787e-01(1.3e-01) & 1.089e+00(4.0e-02) & 4.686e-01(1.0e-01) & 1.859e-01(2.2e-02) & 1.196e+00(5.5e-03) & \cellcolor{gray!50}\textbf{1.197e+00(7.4e-03)} & 1.318e+00(5.1e-03) \\
PSLR & 1.197e+00(1.6e-03) & 3.334e-01(4.8e-02) & \cellcolor{gray!50}\textbf{1.196e+00(1.3e-01)} & 5.584e-01(1.2e-01) & 1.117e+00(4.2e-02) & 5.889e-01(1.3e-01) & 3.116e-01(2.7e-02) & 1.187e+00(3.9e-03) & 1.115e+00(2.4e-02) & 1.310e+00(1.2e-02) \\
PSLO & \cellcolor{gray!50}\textbf{1.199e+00(8.3e-07)} & 1.771e-01(1.1e-01) & 1.014e+00(1.9e-01) & 5.500e-01(1.5e-01) & 1.038e+00(6.4e-02) & 5.425e-01(1.3e-01) & 2.608e-01(4.5e-02) & 1.005e+00(2.1e-01) & 7.207e-01(2.7e-01) & 1.054e+00(2.8e-01) \\
DOMOO & \cellcolor{gray!50}\textbf{1.199e+00(8.7e-06)} & 3.293e-01(8.3e-02) & 1.186e+00(8.6e-02) & 6.968e-01(8.5e-02) & 8.080e-01(2.3e-01) & 5.148e-01(1.0e-01) & 2.869e-01(2.8e-02) & \cellcolor{gray!50}\textbf{1.205e+00(4.2e-03)} & 1.132e+00(6.9e-02) & \cellcolor{gray!50}\textbf{1.327e+00(3.5e-03)} \\
\midrule
ICPSL-MAP & 1.198e+00(2.4e-03) & 1.496e-01(4.7e-02) & 8.693e-01(2.2e-01) & 6.083e-01(1.5e-01) & \cellcolor{gray!50}\textbf{1.122e+00(3.0e-02)} & 5.377e-01(1.2e-01) & 2.222e-01(4.4e-02) & 1.178e+00(6.3e-03) & 1.130e+00(2.5e-02) & 1.304e+00(8.7e-03) \\
ICPSL-CAM & \cellcolor{gray!50}\textbf{1.199e+00(3.6e-04)} & 4.087e-01(3.0e-02) & 9.261e-01(2.0e-01) & 7.639e-01(1.0e-01) & 1.087e+00(6.1e-02) & 5.710e-01(9.4e-02) & 3.116e-01(2.4e-02) & 1.183e+00(5.8e-03) & 1.126e+00(1.8e-02) & 1.301e+00(9.2e-03) \\
ICPSL-MM & \cellcolor{gray!50}\textbf{1.199e+00(3.8e-04)} & \cellcolor{gray!50}\textbf{4.429e-01(2.3e-02)} & 9.818e-01(1.8e-01) & \cellcolor{gray!50}\textbf{7.793e-01(1.0e-01)} & 1.089e+00(5.7e-02) & \cellcolor{gray!50}\textbf{5.932e-01(9.1e-02)} & \cellcolor{gray!50}\textbf{3.168e-01(1.7e-02)} & 1.182e+00(5.8e-03) & 1.118e+00(1.8e-02) & 1.299e+00(9.3e-03) \\
\bottomrule[1pt]
\end{tabular}%
}
\end{table}

\begin{table}[!ht]
\centering
\caption{Mean hypervolume ($\uparrow$) with 95\% confidence intervals using uniform context over 10 independent runs.}
\label{tab:app-hv-uniform}
\renewcommand{\arraystretch}{1.5}
\resizebox{\textwidth}{!}{%
\begin{tabular}{lcccccccccc}
\toprule[1pt]
Algorithm & VLMOP1 & VLMOP2 & VLMOP3 & AR & DR & ARA & ABC & RE23 & RE24 & RE32 \\
\midrule
ParetoFlow & 1.185e+00(9.5e-03) & 1.679e-01(4.3e-02) & 9.592e-01(1.9e-01) & 6.662e-01(1.3e-01) & 1.003e+00(5.5e-02) & 6.411e-01(1.4e-01) & 2.439e-01(2.8e-02) & \cellcolor{gray!50}\textbf{1.190e+00(1.1e-02)} & \cellcolor{gray!50}\textbf{1.169e+00(4.3e-02)} & 1.312e+00(1.0e-02) \\
PSLR & 1.196e+00(2.9e-03) & 3.301e-01(3.7e-02) & 1.043e+00(1.7e-01) & 6.447e-01(1.4e-01) & 1.081e+00(5.3e-02) & 6.002e-01(1.5e-01) & 2.794e-01(3.1e-02) & 1.184e+00(8.6e-03) & 1.089e+00(5.5e-02) & 1.297e+00(1.6e-02) \\
PSLO & 1.196e+00(6.0e-03) & 6.835e-02(2.3e-02) & 1.037e+00(1.7e-01) & 7.567e-01(1.2e-01) & 9.210e-01(9.2e-02) & 5.642e-01(1.3e-01) & 2.773e-01(3.2e-02) & 8.950e-01(2.5e-01) & 7.684e-01(2.7e-01) & 1.045e+00(3.1e-01) \\
DOMOO & \cellcolor{gray!50}\textbf{1.199e+00(3.3e-05)} & 2.162e-01(9.0e-02) & 1.011e+00(1.9e-01) & 5.823e-01(1.3e-01) & 9.521e-01(1.7e-01) & 4.719e-01(1.3e-01) & 2.037e-01(4.6e-02) & 1.114e+00(1.7e-01) & 1.015e+00(2.0e-01) & \cellcolor{gray!50}\textbf{1.330e+00(8.1e-04)} \\
\midrule
ICPSL-MAP & 1.175e+00(3.4e-02) & 1.107e-01(4.1e-02) & 1.060e+00(1.6e-01) & 7.417e-01(1.4e-01) & 1.085e+00(3.7e-02) & 6.496e-01(1.5e-01) & 2.504e-01(3.3e-02) & 1.161e+00(1.3e-02) & 1.104e+00(5.3e-02) & 1.289e+00(2.4e-02) \\
ICPSL-CAM & 1.198e+00(1.2e-03) & 3.182e-01(6.9e-02) & 1.084e+00(1.5e-01) & 7.952e-01(1.3e-01) & 1.087e+00(2.8e-02) & 6.646e-01(1.3e-01) & 2.940e-01(2.3e-02) & 1.172e+00(1.2e-02) & 1.099e+00(5.1e-02) & 1.288e+00(2.1e-02) \\
ICPSL-MM & 1.198e+00(1.3e-03) & \cellcolor{gray!50}\textbf{3.631e-01(5.0e-02)} & \cellcolor{gray!50}\textbf{1.101e+00(1.3e-01)} & \cellcolor{gray!50}\textbf{8.081e-01(1.2e-01)} & \cellcolor{gray!50}\textbf{1.090e+00(2.7e-02)} & \cellcolor{gray!50}\textbf{6.714e-01(1.3e-01)} & \cellcolor{gray!50}\textbf{3.060e-01(2.2e-02)} & 1.172e+00(1.2e-02) & 1.090e+00(5.0e-02) & 1.286e+00(2.1e-02) \\
\bottomrule[1pt]
\end{tabular}%
}
\end{table}

\subsection{Computational Efficiency}

\begin{table}[!ht]
\centering
\caption{Mean training / inference time in seconds ($\downarrow$) using LHS context over 10 independent runs.}
\label{tab:app-time-lhs}
\renewcommand{\arraystretch}{1.5}
\resizebox{\textwidth}{!}{%
\begin{tabular}{lcccccccccc}
\toprule[1pt]
Algorithm & VLMOP1 & VLMOP2 & VLMOP3 & AR & DR & ARA & ABC & RE23 & RE24 & RE32 \\
\midrule
ParetoFlow & 7.60e+00/8.89e-01 & 7.25e+00/8.84e-01 & 7.24e+00/9.33e-01 & 8.47e+00/9.33e-01 & 7.19e+00/8.92e-01 & 7.14e+00/8.84e-01 & 8.63e+00/1.05e+00 & 7.19e+00/8.90e-01 & 7.15e+00/8.85e-01 & 7.25e+00/9.42e-01 \\
PSLR & 2.54e+00/8.53e-04 & 2.64e+00/5.57e-04 & 2.45e+00/6.17e-04 & 2.63e+00/8.67e-04 & 2.48e+00/1.23e-03 & 2.47e+00/1.36e-03 & 2.44e+00/1.28e-03 & 2.49e+00/8.12e-04 & 2.35e+00/1.25e-03 & 2.24e+00/1.84e-02 \\
PSLO & 3.07e+00/4.62e-04 & 3.36e+00/7.12e-04 & 3.12e+00/4.81e-04 & 4.58e+00/1.06e-03 & 4.83e+00/8.63e-04 & 4.62e+00/7.92e-04 & 5.20e+00/1.08e-03 & 4.95e+00/6.29e-04 & 4.19e+00/4.99e-04 & 5.23e+00/9.04e-04 \\
DOMOO & 6.53e+00/1.98e-04 & 5.63e+00/1.90e-04 & 7.89e+00/1.83e-04 & 7.13e+00/2.04e-04 & 8.69e+00/1.94e-04 & 8.83e+00/2.18e-04 & 1.39e+01/2.22e-04 & 9.63e+00/2.11e-04 & 9.22e+00/1.82e-04 & 1.30e+01/1.90e-04 \\
\midrule
ICPSL-MAP & \cellcolor{gray!50}\textbf{4.91e-02}/1.17e-03 & \cellcolor{gray!50}\textbf{3.48e-03}/1.84e-03 & \cellcolor{gray!50}\textbf{3.51e-03}/1.42e-03 & \cellcolor{gray!50}\textbf{3.21e-03}/1.38e-03 & \cellcolor{gray!50}\textbf{3.53e-03}/1.41e-03 & \cellcolor{gray!50}\textbf{6.42e-02}/1.38e-03 & \cellcolor{gray!50}\textbf{3.47e-03}/1.42e-03 & \cellcolor{gray!50}\textbf{4.75e-02}/2.25e-03 & \cellcolor{gray!50}\textbf{3.60e-03}/2.81e-03 & \cellcolor{gray!50}\textbf{3.48e-03}/2.30e-03 \\
ICPSL-CAM & \cellcolor{gray!50}\textbf{4.91e-02}/6.74e-04 & \cellcolor{gray!50}\textbf{3.48e-03}/6.62e-04 & \cellcolor{gray!50}\textbf{3.51e-03}/6.66e-04 & \cellcolor{gray!50}\textbf{3.21e-03}/6.29e-04 & \cellcolor{gray!50}\textbf{3.53e-03}/6.55e-04 & \cellcolor{gray!50}\textbf{6.42e-02}/6.77e-04 & \cellcolor{gray!50}\textbf{3.47e-03}/6.55e-04 & \cellcolor{gray!50}\textbf{4.75e-02}/6.55e-04 & \cellcolor{gray!50}\textbf{3.60e-03}/6.53e-04 & \cellcolor{gray!50}\textbf{3.48e-03}/1.62e-03 \\
ICPSL-MM & \cellcolor{gray!50}\textbf{4.91e-02}/3.38e-03 & \cellcolor{gray!50}\textbf{3.48e-03}/1.91e-03 & \cellcolor{gray!50}\textbf{3.51e-03}/1.71e-03 & \cellcolor{gray!50}\textbf{3.21e-03}/1.73e-03 & \cellcolor{gray!50}\textbf{3.53e-03}/1.73e-03 & \cellcolor{gray!50}\textbf{6.42e-02}/3.12e-03 & \cellcolor{gray!50}\textbf{3.47e-03}/1.74e-03 & \cellcolor{gray!50}\textbf{4.75e-02}/2.03e-03 & \cellcolor{gray!50}\textbf{3.60e-03}/1.69e-03 & \cellcolor{gray!50}\textbf{3.48e-03}/2.10e-03 \\
\bottomrule[1pt]
\end{tabular}%
}
\end{table}

\begin{table}[!ht]
\centering
\caption{Mean training / inference time in seconds ($\downarrow$) using Sobol context over 10 independent runs.}
\label{tab:app-time-sobol}
\renewcommand{\arraystretch}{1.5}
\resizebox{\textwidth}{!}{%
\begin{tabular}{lcccccccccc}
\toprule[1pt]
Algorithm & VLMOP1 & VLMOP2 & VLMOP3 & AR & DR & ARA & ABC & RE23 & RE24 & RE32 \\
\midrule
ParetoFlow & 7.75e+00/9.20e-01 & 8.40e+00/9.00e-01 & 7.28e+00/9.39e-01 & 7.30e+00/8.97e-01 & 7.16e+00/8.86e-01 & 7.27e+00/8.98e-01 & 7.77e+00/1.02e+00 & 7.29e+00/9.17e-01 & 8.33e+00/9.01e-01 & 8.36e+00/9.68e-01 \\
PSLR & 2.49e+00/1.32e-03 & 2.73e+00/4.17e-04 & 2.43e+00/6.01e-04 & 2.57e+00/9.40e-04 & 2.38e+00/9.86e-04 & 2.31e+00/1.32e-03 & 2.57e+00/1.09e-03 & 2.12e+00/9.25e-04 & 2.41e+00/8.07e-04 & 2.30e+00/9.56e-03 \\
PSLO & 3.24e+00/9.60e-04 & 3.51e+00/2.68e-04 & 3.73e+00/8.59e-04 & 3.93e+00/5.54e-04 & 4.91e+00/8.45e-04 & 4.48e+00/5.25e-04 & 5.59e+00/5.07e-04 & 5.23e+00/4.74e-04 & 5.24e+00/5.66e-04 & 5.56e+00/6.58e-04 \\
DOMOO & 6.91e+00/2.04e-04 & 5.80e+00/2.01e-04 & 8.99e+00/2.00e-04 & 9.00e+00/2.29e-04 & 1.01e+01/2.05e-04 & 8.11e+00/2.21e-04 & 1.36e+01/1.83e-04 & 9.15e+00/3.54e-04 & 9.85e+00/1.99e-04 & 1.30e+01/1.85e-04 \\
\midrule
ICPSL-MAP & \cellcolor{gray!50}\textbf{3.24e-03}/1.22e-03 & \cellcolor{gray!50}\textbf{3.62e-03}/1.85e-03 & \cellcolor{gray!50}\textbf{3.25e-03}/1.38e-03 & \cellcolor{gray!50}\textbf{3.32e-02}/1.43e-03 & \cellcolor{gray!50}\textbf{4.66e-02}/1.48e-03 & \cellcolor{gray!50}\textbf{3.51e-03}/1.43e-03 & \cellcolor{gray!50}\textbf{4.81e-02}/1.43e-03 & \cellcolor{gray!50}\textbf{3.59e-03}/2.25e-03 & \cellcolor{gray!50}\textbf{3.52e-03}/1.44e-03 & \cellcolor{gray!50}\textbf{3.49e-03}/2.22e-03 \\
ICPSL-CAM & \cellcolor{gray!50}\textbf{3.24e-03}/6.60e-04 & \cellcolor{gray!50}\textbf{3.62e-03}/2.79e-03 & \cellcolor{gray!50}\textbf{3.25e-03}/6.39e-04 & \cellcolor{gray!50}\textbf{3.32e-02}/6.71e-04 & \cellcolor{gray!50}\textbf{4.66e-02}/6.31e-04 & \cellcolor{gray!50}\textbf{3.51e-03}/6.81e-04 & \cellcolor{gray!50}\textbf{4.81e-02}/6.67e-04 & \cellcolor{gray!50}\textbf{3.59e-03}/6.58e-04 & \cellcolor{gray!50}\textbf{3.52e-03}/6.36e-04 & \cellcolor{gray!50}\textbf{3.49e-03}/6.77e-04 \\
ICPSL-MM & \cellcolor{gray!50}\textbf{3.24e-03}/2.25e-03 & \cellcolor{gray!50}\textbf{3.62e-03}/1.86e-03 & \cellcolor{gray!50}\textbf{3.25e-03}/1.74e-03 & \cellcolor{gray!50}\textbf{3.32e-02}/1.73e-03 & \cellcolor{gray!50}\textbf{4.66e-02}/4.48e-03 & \cellcolor{gray!50}\textbf{3.51e-03}/1.77e-03 & \cellcolor{gray!50}\textbf{4.81e-02}/1.73e-03 & \cellcolor{gray!50}\textbf{3.59e-03}/2.01e-03 & \cellcolor{gray!50}\textbf{3.52e-03}/1.90e-03 & \cellcolor{gray!50}\textbf{3.49e-03}/3.67e-03 \\
\bottomrule[1pt]
\end{tabular}%
}
\end{table}

\begin{table}[!ht]
\centering
\caption{Mean training / inference time in seconds ($\downarrow$) using uniform context over 10 independent runs.}
\label{tab:app-time-uniform}
\renewcommand{\arraystretch}{1.5}
\resizebox{\textwidth}{!}{%
\begin{tabular}{lcccccccccc}
\toprule[1pt]
Algorithm & VLMOP1 & VLMOP2 & VLMOP3 & AR & DR & ARA & ABC & RE23 & RE24 & RE32 \\
\midrule
ParetoFlow & 7.71e+00/9.27e-01 & 8.57e+00/9.36e-01 & 8.40e+00/1.01e+00 & 7.23e+00/9.14e-01 & 7.32e+00/9.24e-01 & 7.24e+00/9.10e-01 & 8.63e+00/1.01e+00 & 7.16e+00/8.82e-01 & 7.22e+00/8.81e-01 & 8.51e+00/9.53e-01 \\
PSLR & 2.57e+00/7.19e-04 & 2.75e+00/7.95e-04 & 2.44e+00/1.07e-02 & 2.59e+00/1.41e-03 & 2.53e+00/1.21e-03 & 2.43e+00/1.20e-03 & 2.55e+00/1.24e-03 & 2.50e+00/1.24e-03 & 2.37e+00/9.88e-04 & 2.33e+00/1.28e-03 \\
PSLO & 2.97e+00/7.03e-04 & 2.74e+00/4.93e-04 & 3.27e+00/8.91e-04 & 4.80e+00/7.41e-04 & 4.48e+00/8.14e-04 & 4.93e+00/4.09e-04 & 4.49e+00/4.96e-04 & 4.50e+00/7.83e-04 & 4.42e+00/7.95e-04 & 5.15e+00/5.95e-04 \\
DOMOO & 5.99e+00/2.56e-04 & 5.89e+00/2.02e-04 & 1.06e+01/1.96e-04 & 6.97e+00/1.85e-04 & 9.57e+00/1.93e-04 & 9.29e+00/1.79e-04 & 1.37e+01/1.83e-04 & 9.16e+00/2.02e-04 & 9.28e+00/1.80e-04 & 1.41e+01/2.16e-04 \\
\midrule
ICPSL-MAP & \cellcolor{gray!50}\textbf{4.78e-02}/1.12e-03 & \cellcolor{gray!50}\textbf{3.65e-03}/1.78e-03 & \cellcolor{gray!50}\textbf{3.63e-03}/1.40e-03 & \cellcolor{gray!50}\textbf{3.31e-03}/1.44e-03 & \cellcolor{gray!50}\textbf{3.51e-03}/1.42e-03 & \cellcolor{gray!50}\textbf{4.81e-02}/1.47e-03 & \cellcolor{gray!50}\textbf{3.23e-03}/1.42e-03 & \cellcolor{gray!50}\textbf{4.77e-02}/3.52e-03 & \cellcolor{gray!50}\textbf{3.59e-03}/1.38e-03 & \cellcolor{gray!50}\textbf{3.52e-03}/2.26e-03 \\
ICPSL-CAM & \cellcolor{gray!50}\textbf{4.78e-02}/6.60e-04 & \cellcolor{gray!50}\textbf{3.65e-03}/6.85e-04 & \cellcolor{gray!50}\textbf{3.63e-03}/6.43e-04 & \cellcolor{gray!50}\textbf{3.31e-03}/6.71e-04 & \cellcolor{gray!50}\textbf{3.51e-03}/6.19e-04 & \cellcolor{gray!50}\textbf{4.81e-02}/6.37e-04 & \cellcolor{gray!50}\textbf{3.23e-03}/6.57e-04 & \cellcolor{gray!50}\textbf{4.77e-02}/6.17e-04 & \cellcolor{gray!50}\textbf{3.59e-03}/6.42e-04 & \cellcolor{gray!50}\textbf{3.52e-03}/6.62e-04 \\
ICPSL-MM & \cellcolor{gray!50}\textbf{4.78e-02}/2.22e-03 & \cellcolor{gray!50}\textbf{3.65e-03}/2.84e-03 & \cellcolor{gray!50}\textbf{3.63e-03}/1.73e-03 & \cellcolor{gray!50}\textbf{3.31e-03}/1.73e-03 & \cellcolor{gray!50}\textbf{3.51e-03}/1.75e-03 & \cellcolor{gray!50}\textbf{4.81e-02}/1.70e-03 & \cellcolor{gray!50}\textbf{3.23e-03}/1.73e-03 & \cellcolor{gray!50}\textbf{4.77e-02}/5.71e-03 & \cellcolor{gray!50}\textbf{3.59e-03}/1.73e-03 & \cellcolor{gray!50}\textbf{3.52e-03}/2.00e-03 \\
\bottomrule[1pt]
\end{tabular}%
}
\end{table}

\section{Detailed Results for Additional Analyses}
\label{sec:detailed-additional-results}

\subsection{Evolution of Solution-Model Quality}
\label{sec:evolution-solution-model-quality-results}

\begin{figure}[!h]
\centering
\includegraphics[width=\textwidth]{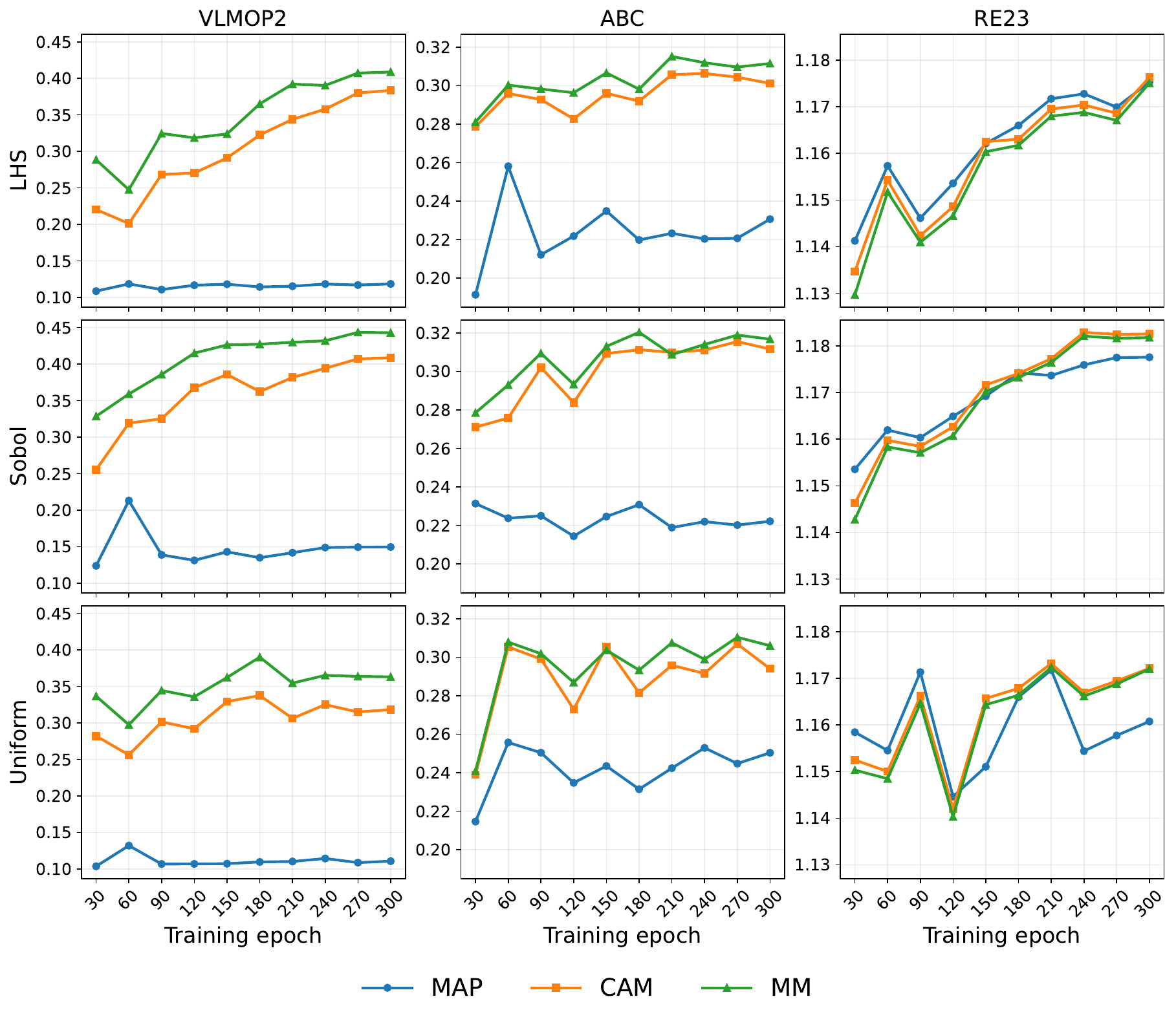}
\caption{Evolution of hypervolume (higher is better) for ICPSL-MAP, ICPSL-CAM, and ICPSL-MM throughout pretraining under LHS, Sobol, and uniform context distributions.}
\label{fig:exp-hv-vs-epoch}
\end{figure}

\clearpage

\subsection{Sensitivity to Context Size}
\label{sec:sensitivity-context-size-results}

\begin{figure}[!h]
\centering
\includegraphics[width=\textwidth]{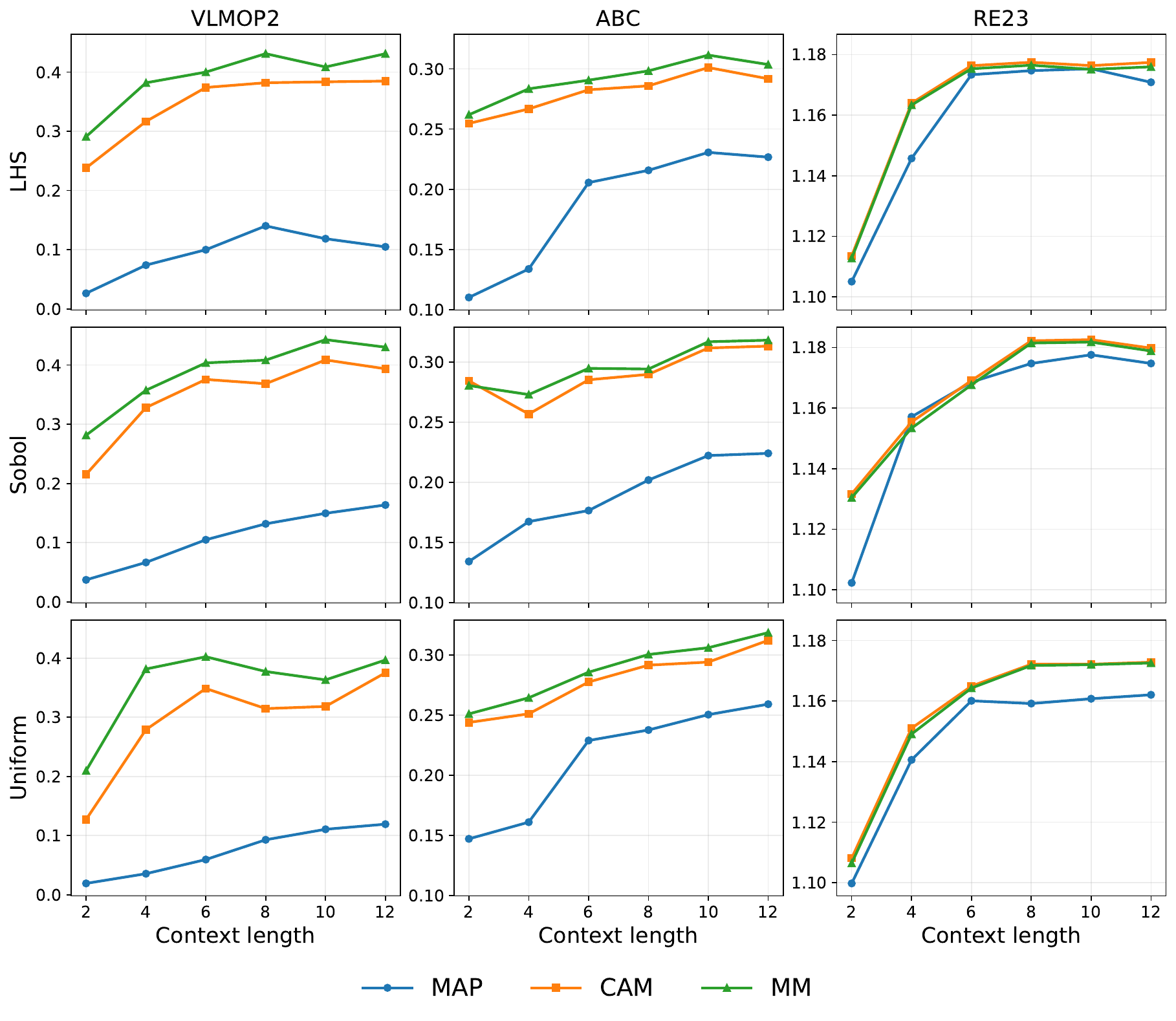}
\caption{Hypervolume (higher is better) of ICPSL-MAP, ICPSL-CAM, and ICPSL-MM with varying context sizes under LHS, Sobol, and uniform context distributions.}
\label{fig:exp-hv-vs-context-size}
\end{figure}

\clearpage

\subsection{Generalization to OOD Dimensions}
\label{sec:ood-dimension-results}

\begin{table}[!ht]
\centering
\caption{Mean hypervolume ($\uparrow$) with 95\% confidence intervals on OOD-dimensional benchmarks using LHS context over 10 independent runs.}
\label{tab:app-ood-hv-lhs}
\renewcommand{\arraystretch}{1.5}
\resizebox{\textwidth}{!}{%
\begin{tabular}{lcccc}
\toprule[1pt]
Algorithm & AR (5d2m) & DR (5d2m) & ARA (5d2m) & DTLZ4 (4d4m) \\
\midrule
ParetoFlow & 2.399e-01(2.5e-02) & 8.306e-01(4.3e-02) & 1.687e-01(1.3e-02) & \cellcolor{gray!50}\textbf{6.356e-01(1.1e-01)} \\
PSLR & 2.852e-01(4.9e-02) & 9.703e-01(6.8e-02) & 2.367e-01(4.0e-02) & 4.873e-01(8.3e-02) \\
PSLO & 2.397e-01(3.0e-02) & 6.403e-01(1.0e-01) & 1.441e-01(3.1e-02) & 3.958e-01(5.5e-03) \\
DOMOO & 2.193e-01(6.2e-02) & 7.125e-01(1.3e-01) & 1.312e-01(2.3e-02) & 3.922e-01(6.0e-02) \\
\midrule
ICPSL-MAP & 3.065e-01(3.8e-02) & 9.214e-01(3.7e-02) & 1.736e-01(3.3e-02) & 4.538e-01(4.7e-02) \\
ICPSL-CAM & 4.945e-01(7.9e-02) & 9.875e-01(4.2e-02) & 4.593e-01(9.7e-02) & 4.032e-01(3.9e-04) \\
ICPSL-MM & \cellcolor{gray!50}\textbf{5.574e-01(7.5e-02)} & \cellcolor{gray!50}\textbf{1.025e+00(3.4e-02)} & \cellcolor{gray!50}\textbf{5.457e-01(8.1e-02)} & 4.030e-01(5.7e-04) \\
\bottomrule[1pt]
\end{tabular}%
}
\end{table}

\begin{table}[!ht]
\centering
\caption{Mean hypervolume ($\uparrow$) with 95\% confidence intervals on OOD-dimensional benchmarks using Sobol context over 10 independent runs.}
\label{tab:app-ood-hv-sobol}
\renewcommand{\arraystretch}{1.5}
\resizebox{\textwidth}{!}{%
\begin{tabular}{lcccc}
\toprule[1pt]
Algorithm & AR (5d2m) & DR (5d2m) & ARA (5d2m) & DTLZ4 (4d4m) \\
\midrule
ParetoFlow & 2.489e-01(2.2e-02) & 8.671e-01(3.9e-02) & 1.871e-01(2.3e-02) & \cellcolor{gray!50}\textbf{6.481e-01(8.8e-02)} \\
PSLR & 3.081e-01(3.2e-02) & 9.528e-01(5.0e-02) & 2.803e-01(5.1e-02) & 4.199e-01(3.1e-02) \\
PSLO & 2.404e-01(5.6e-02) & 7.160e-01(1.4e-01) & 1.798e-01(3.1e-02) & 3.535e-01(6.9e-02) \\
DOMOO & 2.360e-01(5.2e-02) & 5.169e-01(1.6e-01) & 1.257e-01(1.6e-02) & 3.537e-01(6.9e-02) \\
\midrule
ICPSL-MAP & 3.455e-01(6.4e-02) & 9.219e-01(4.9e-02) & 2.484e-01(5.4e-02) & 4.349e-01(5.9e-02) \\
ICPSL-CAM & 5.483e-01(9.0e-02) & 9.883e-01(5.2e-02) & 4.659e-01(5.1e-02) & 4.031e-01(4.6e-04) \\
ICPSL-MM & \cellcolor{gray!50}\textbf{6.151e-01(9.8e-02)} & \cellcolor{gray!50}\textbf{9.897e-01(4.4e-02)} & \cellcolor{gray!50}\textbf{5.549e-01(5.7e-02)} & 4.029e-01(6.8e-04) \\
\bottomrule[1pt]
\end{tabular}%
}
\end{table}

\begin{table}[!ht]
\centering
\caption{Mean hypervolume ($\uparrow$) with 95\% confidence intervals on OOD-dimensional benchmarks using uniform context over 10 independent runs.}
\label{tab:app-ood-hv-uniform}
\renewcommand{\arraystretch}{1.5}
\resizebox{\textwidth}{!}{%
\begin{tabular}{lcccc}
\toprule[1pt]
Algorithm & AR (5d2m) & DR (5d2m) & ARA (5d2m) & DTLZ4 (4d4m) \\
\midrule
ParetoFlow & 2.969e-01(4.8e-02) & 8.684e-01(3.1e-02) & 2.104e-01(4.2e-02) & \cellcolor{gray!50}\textbf{6.231e-01(1.0e-01)} \\
PSLR & 3.537e-01(7.5e-02) & 9.302e-01(6.1e-02) & 2.809e-01(5.6e-02) & 4.888e-01(8.9e-02) \\
PSLO & 2.365e-01(4.0e-02) & 6.416e-01(1.1e-01) & 1.562e-01(3.5e-02) & 4.017e-01(1.6e-02) \\
DOMOO & 1.441e-01(3.5e-02) & 5.239e-01(1.7e-01) & 1.310e-01(4.5e-02) & 3.711e-01(7.6e-02) \\
\midrule
ICPSL-MAP & 3.364e-01(5.9e-02) & 9.235e-01(5.3e-02) & 2.185e-01(6.0e-02) & 4.292e-01(2.8e-02) \\
ICPSL-CAM & 5.001e-01(5.1e-02) & 9.746e-01(4.9e-02) & 4.214e-01(5.7e-02) & 4.031e-01(5.3e-04) \\
ICPSL-MM & \cellcolor{gray!50}\textbf{5.625e-01(6.1e-02)} & \cellcolor{gray!50}\textbf{9.834e-01(4.6e-02)} & \cellcolor{gray!50}\textbf{5.084e-01(6.5e-02)} & 4.031e-01(5.2e-04) \\
\bottomrule[1pt]
\end{tabular}%
}
\end{table}

\begin{table}[!ht]
\centering
\caption{Mean training / inference time in seconds ($\downarrow$) on OOD-dimensional benchmarks using LHS context over 10 independent runs.}
\label{tab:app-ood-time-lhs}
\renewcommand{\arraystretch}{1.5}
\resizebox{\textwidth}{!}{%
\begin{tabular}{lcccc}
\toprule[1pt]
Algorithm & AR (5d2m) & DR (5d2m) & ARA (5d2m) & DTLZ4 (4d4m) \\
\midrule
ParetoFlow & 9.71e+00/5.56e+00 & 9.30e+00/5.27e+00 & 8.94e+00/4.64e+00 & 7.32e+00/9.88e-01 \\
PSLR & 4.77e-01/1.99e-04 & 5.65e-01/1.71e-04 & 5.44e-01/1.81e-04 & 6.26e-01/1.93e-04 \\
PSLO & 1.58e+00/1.76e-04 & 1.99e+00/1.78e-04 & 2.02e+00/1.87e-04 & 2.63e+00/1.74e-04 \\
DOMOO & 3.76e+00/1.63e-04 & 4.58e+00/1.61e-04 & 4.91e+00/1.73e-04 & 9.99e+00/1.64e-04 \\
\midrule
ICPSL-MAP & \cellcolor{gray!50}\textbf{3.28e-03}/2.91e-03 & \cellcolor{gray!50}\textbf{3.22e-02}/5.67e-03 & \cellcolor{gray!50}\textbf{3.73e-03}/2.78e-03 & \cellcolor{gray!50}\textbf{4.67e-02}/2.18e-03 \\
ICPSL-CAM & \cellcolor{gray!50}\textbf{3.28e-03}/6.29e-04 & \cellcolor{gray!50}\textbf{3.22e-02}/6.27e-04 & \cellcolor{gray!50}\textbf{3.73e-03}/6.68e-04 & \cellcolor{gray!50}\textbf{4.67e-02}/6.63e-04 \\
ICPSL-MM & \cellcolor{gray!50}\textbf{3.28e-03}/2.33e-03 & \cellcolor{gray!50}\textbf{3.22e-02}/3.45e-03 & \cellcolor{gray!50}\textbf{3.73e-03}/2.37e-03 & \cellcolor{gray!50}\textbf{4.67e-02}/2.02e-03 \\
\bottomrule[1pt]
\end{tabular}%
}
\end{table}

\begin{table}[!ht]
\centering
\caption{Mean training / inference time in seconds ($\downarrow$) on OOD-dimensional benchmarks using Sobol context over 10 independent runs.}
\label{tab:app-ood-time-sobol}
\renewcommand{\arraystretch}{1.5}
\resizebox{\textwidth}{!}{%
\begin{tabular}{lcccc}
\toprule[1pt]
Algorithm & AR (5d2m) & DR (5d2m) & ARA (5d2m) & DTLZ4 (4d4m) \\
\midrule
ParetoFlow & 8.39e+00/3.00e+00 & 9.26e+00/4.92e+00 & 9.10e+00/5.37e+00 & 9.50e+00/5.40e+00 \\
PSLR & 5.40e-01/1.72e-04 & 5.32e-01/1.84e-04 & 5.51e-01/1.78e-04 & 5.09e-01/1.92e-04 \\
PSLO & 1.78e+00/2.24e-04 & 1.87e+00/1.76e-04 & 1.83e+00/1.85e-04 & 2.56e+00/2.67e-04 \\
DOMOO & 4.83e+00/1.96e-04 & 4.68e+00/1.65e-04 & 4.79e+00/1.63e-04 & 1.06e+01/1.71e-04 \\
\midrule
ICPSL-MAP & \cellcolor{gray!50}\textbf{3.77e-03}/6.09e-03 & \cellcolor{gray!50}\textbf{3.29e-03}/2.84e-03 & \cellcolor{gray!50}\textbf{6.22e-03}/2.93e-03 & \cellcolor{gray!50}\textbf{3.61e-03}/2.22e-03 \\
ICPSL-CAM & \cellcolor{gray!50}\textbf{3.77e-03}/5.78e-03 & \cellcolor{gray!50}\textbf{3.29e-03}/6.56e-04 & \cellcolor{gray!50}\textbf{6.22e-03}/3.19e-03 & \cellcolor{gray!50}\textbf{3.61e-03}/6.45e-04 \\
ICPSL-MM & \cellcolor{gray!50}\textbf{3.77e-03}/4.20e-03 & \cellcolor{gray!50}\textbf{3.29e-03}/2.26e-03 & \cellcolor{gray!50}\textbf{6.22e-03}/2.25e-03 & \cellcolor{gray!50}\textbf{3.61e-03}/1.98e-03 \\
\bottomrule[1pt]
\end{tabular}%
}
\end{table}

\begin{table}[!ht]
\centering
\caption{Mean training / inference time in seconds ($\downarrow$) on OOD-dimensional benchmarks using uniform context over 10 independent runs.}
\label{tab:app-ood-time-uniform}
\renewcommand{\arraystretch}{1.5}
\resizebox{\textwidth}{!}{%
\begin{tabular}{lcccc}
\toprule[1pt]
Algorithm & AR (5d2m) & DR (5d2m) & ARA (5d2m) & DTLZ4 (4d4m) \\
\midrule
ParetoFlow & 9.29e+00/3.10e+00 & 9.16e+00/5.27e+00 & 9.46e+00/5.46e+00 & 9.70e+00/5.37e+00 \\
PSLR & 4.59e-01/1.75e-04 & 5.30e-01/1.82e-04 & 5.38e-01/1.84e-04 & 5.16e-01/1.84e-04 \\
PSLO & 2.01e+00/1.96e-04 & 1.97e+00/1.76e-04 & 1.88e+00/1.79e-04 & 2.54e+00/1.77e-04 \\
DOMOO & 4.63e+00/1.65e-04 & 4.68e+00/1.89e-04 & 4.95e+00/1.59e-04 & 9.55e+00/1.82e-04 \\
\midrule
ICPSL-MAP & \cellcolor{gray!50}\textbf{3.17e-03}/2.95e-03 & \cellcolor{gray!50}\textbf{3.08e-03}/2.86e-03 & \cellcolor{gray!50}\textbf{3.33e-02}/2.84e-03 & \cellcolor{gray!50}\textbf{3.53e-03}/2.24e-03 \\
ICPSL-CAM & \cellcolor{gray!50}\textbf{3.17e-03}/6.40e-04 & \cellcolor{gray!50}\textbf{3.08e-03}/6.72e-04 & \cellcolor{gray!50}\textbf{3.33e-02}/6.38e-04 & \cellcolor{gray!50}\textbf{3.53e-03}/6.39e-04 \\
ICPSL-MM & \cellcolor{gray!50}\textbf{3.17e-03}/2.25e-03 & \cellcolor{gray!50}\textbf{3.08e-03}/2.25e-03 & \cellcolor{gray!50}\textbf{3.33e-02}/2.27e-03 & \cellcolor{gray!50}\textbf{3.53e-03}/1.99e-03 \\
\bottomrule[1pt]
\end{tabular}%
}
\end{table}